\pdfoutput=1
\documentclass[11pt]{article}

\usepackage[preprint]{acl}

\usepackage{times}
\usepackage{latexsym}

\usepackage[T1]{fontenc}
\usepackage[utf8]{inputenc}

\usepackage{microtype}

\usepackage{inconsolata}

\usepackage{hyperref}
\usepackage{graphicx}
\usepackage{xspace}
\usepackage{paralist}
\usepackage{wrapfig}
\usepackage{booktabs}
\usepackage{multirow}
\usepackage{makecell}
\usepackage{array}
\usepackage{amsmath}
\usepackage{amsfonts}
\usepackage{amssymb}
\usepackage{pifont}
\usepackage{mathrsfs}
\usepackage{algorithm,algpseudocode}
\usepackage{mdwlist}
\usepackage{enumitem}
\usepackage{xcolor}
\usepackage{dsfont}
\usepackage{amsthm}
\usepackage{bm}
\usepackage{float}
\usepackage{caption}
\usepackage{subcaption}
\usepackage{tcolorbox}
\usepackage[normalem]{ulem}
\usepackage{tabularx}
\usepackage{colortbl}
\usepackage{CJKutf8}
\usepackage{CJKfntef}
\usepackage{listings}

\definecolor{lightgray}{gray}{0.98}
\definecolor{mypink}{HTML}{FA759E}
\definecolor{myblue}{HTML}{798FF2}
\definecolor{myyellow}{HTML}{FFC96E}

\usepackage[compact]{titlesec}
\usepackage{sidecap}
\titlespacing*{\section}{0pt}{*0.15}{*0.15}
\titlespacing*{\subsection}{0pt}{*0.15}{*0.15}
\titlespacing*{\subsubsection}{0pt}{*0.1}{*0.1}
\renewcommand{\paragraph}[1]{\noindent\textbf{#1}}

\title{Multi-agent KTO: Reinforcing Strategic Interactions of Large Language Model in Language Game}
\author{
  Rong Ye\textsuperscript{\rm 1,2}\protect\footnotemark[1]~, 
  Yongxin Zhang\textsuperscript{\rm 2}\protect\footnotemark[1]~, 
  Yikai Zhang\textsuperscript{\rm 1,2}\protect\footnotemark[2]~, 
  Haoyu Kuang\textsuperscript{\rm 1,2}\protect\footnotemark[2]~,\\
  \textbf{Peng Sun\textsuperscript{\rm 2},
          Zhongyu Wei\textsuperscript{\rm 1}} \\
  \textsuperscript{\rm 1} Fudan University\quad
  \textsuperscript{\rm 2} Bytedance Research \\
  \texttt{\{yerong, zhangyongxin.yx, wanhesong\}@bytedance.com}\\
  \texttt{\{hykuang23, ykzhang22\}@m.fudan.edu.cn}, \texttt{zywei@fudan.edu.cn}
}

\begin{document}
\maketitle
\begin{abstract}
Achieving Artificial General Intelligence (AGI) requires AI agents that can not only make strategic decisions but also engage in flexible and meaningful communication.
Inspired by Wittgenstein's language game theory, we propose that language agents can learn through in-context interaction rather than traditional multi-stage frameworks that separate decision-making from language expression. 
Using \textit{Werewolf}, a social deduction game that tests language understanding, strategic interaction, and adaptability, as a test bed, we develop the Multi-agent Kahneman-Tversky's Optimization (MaKTO). 
MaKTO engages diverse models in extensive gameplay to generate unpaired desirable and unacceptable responses, then employs KTO to refine the model's decision-making process. 
In 9-player \textit{Werewolf} games, MaKTO achieves a 61\% average win rate across various models, outperforming GPT-4o and two-stage RL agents by relative improvements of 23.0\% and 10.9\%, respectively.
Notably, MaKTO also demonstrates human-like performance, winning 60\% against expert players and showing only 48.9\% detectability in Turing-style blind tests. 
% These results showcase MaKTO's superior decision-making, strategic adaptation, and natural language generation in complex social deduction games.
% \footnote{Code and data will be available at \url{https://reneeye.github.io/MaKTO.html}.}
\end{abstract}

\section{Introduction}
\label{sec:intro}

\begin{figure}
    \centering
    \includegraphics[width=1.02\linewidth]{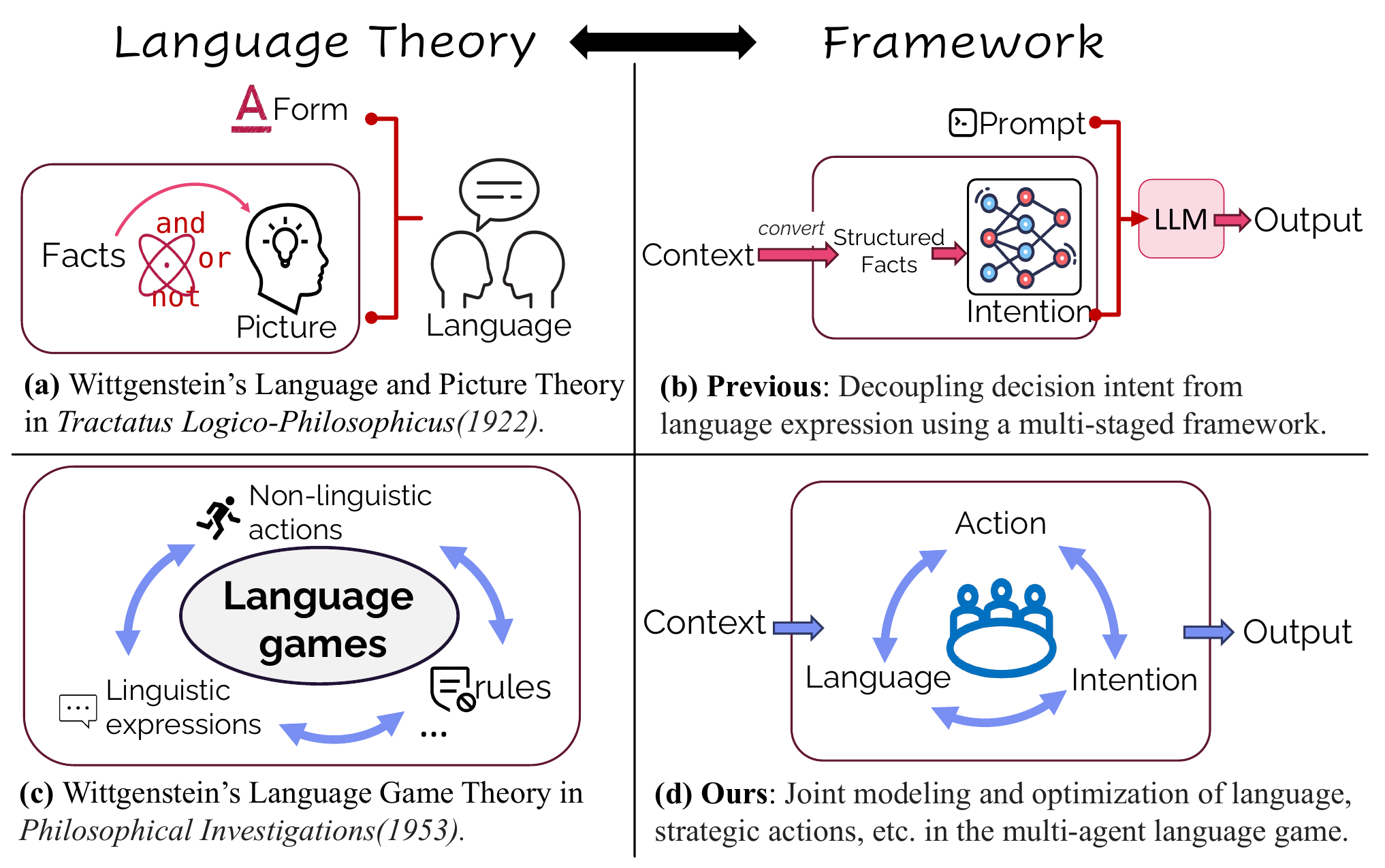}
    \caption{\textbf{Language Theory and AI Architecture: Traditional vs. Language Game Models}.\textit{(a) and (b)}: Multi-staged framework that separates language and decisions. \textit{(c) and (d)}: our proposed framework inspired by Wittgenstein's language game theory, integrating language, actions, and intentions in a multi-agent game.}
    \label{fig:theory_frame}
\end{figure}

Building language agents capable of both decision-making and dialogue represents a crucial pathway toward Artificial General Intelligence (AGI)~\cite{wooldridge1995intelligent,reed2022generalist,salmon2023managing, sumers2023cognitive}. 
This pursuit necessitates a deep understanding of the intrinsic relationship between language and intelligence. 
Wittgenstein's Language Game Theory offers a profound insight: 
meaning emerges from linguistic and non-linguistic interactions, regulated by social rules, forming \textit{language games}~(Fig.\ref{fig:theory_frame}c)~\cite{wittgenstein1953philosophical,kopytko2007philosophy}.
% language derives its meaning through the interplay of linguistic expressions and non-linguistic actions, governed by rules that emerge from our shared forms of life, where all these elements are interwoven into what he terms \textit{language games}~(Fig.~\ref{fig:theory_frame}(c))~\cite{wittgenstein1953philosophical,kopytko2007philosophy}.
% In this language theory framework, language itself constitutes a form of action, with meaning emerging through practical use. 
This contrasts with his earlier Tractatus view of language as logical reality-mapping (Fig.\ref{fig:theory_frame}a).
% This perspective represents a significant departure from his earlier work in \textit{Tractatus Logico-Philosophicus}~\cite{wittgenstein-1922}, where he viewed language as a rigid logical picture of reality~(Fig. \ref{fig:theory_frame}(a)). 
This theoretical perspective points to the value of grounding AI development in practical language use and authentic interactive contexts~\cite{harre1988wittgenstein,wang2016learning,schaul2024boundless,cuskley2024limitations}.

Social deduction games serve as excellent testbeds for validating these theoretical principles, with self-contained language-game environments that test multiple capabilities including linguistic skills, strategic social interaction, and adaptability~\cite{schlangen2023dialogue,wu2023jubensha,trevisan2024measuring,light2023avalonbench,chi2024amongagents,wang2024battleagentbench}. These games provide quantifiable metrics like completion and win rates, enabling direct comparisons between human and AI performance. 
The \textit{Werewolf} game\footnote{Also known as \textit{Mafia}, \url{https://en.wikipedia.org/wiki/Mafia_(party_game)}, the detailed game introduction is in Appendix~\ref{sec:app:game_rule}.}, as a popular and typical social deduction game, exemplifies these characteristics, making it a challenging testbed for AI agent~\cite{xu2023exploring,bailis2024werewolf,xu2023language,wu2024enhance,du2024helmsman}.

% Social deduction games emerge as an ideal testbed for validating these theoretical principles. These games create self-contained language-game ecosystems that serve as excellent research and testing platforms~\cite{xu2023exploring,schlangen2023dialogue,wu2023jubensha,trevisan2024measuring,light2023avalonbench,chi2024amongagents,wang2024battleagentbench}. 
% They require multiple capabilities: 
% \textit{1) fundamental linguistic skills}, including natural language understanding and generation, and hidden information inference; 
% \textit{2) strategic social interaction}, including consistent role-playing and strategic planning for competition and cooperation; and 
% \textit{3) adaptability}, including strategy adjustment and error correction during gameplay. 
% % This comprehensive set of requirements provides an effective framework for evaluating both the pragmatic capabilities and cognitive flexibility of AI systems.
% Furthermore, these games offer quantifiable metrics, such as completion and win rates, enabling systematic comparison between human and AI performance.
% The \textit{Werewolf} game\footnote{Also known as \textit{Mafia}, \url{https://en.wikipedia.org/wiki/Mafia_(party_game)}}~(game introduction is in Appendix~\ref{app:game_rules}), as a popular and typical social deduction game, exemplifies these characteristics, making it a challenging testbed for AI agent~\cite{xu2023exploring,bailis2024werewolf,xu2023language,wu2024enhance,du2024helmsman}.

Current approaches to building AI agents often decouple language processing from decision-making~\cite{he2018decoupling,wu2024enhance,taniguchi2024generative}, echoing the limitations of the picture theory in Fig.\ref{fig:theory_frame}a --- where complex social dynamics are oversimplified into rigid representations.
For example, \citet{wu2024enhance}, as illustrated in Fig.~\ref{fig:theory_frame}b, applied RL policy for decision-making or intention generation, then followed by a large language model~(LLM) for dialogue generation in \textit{Werewolf} game. However, they compressed the language input into structured facts, limiting generalization and cross-environment strategy transfer.
% while hindering multi-agent cooperation.
% for the decision module, which limited the system's generalization capability in decision-making. This multi-staged framework also makes it difficult to transfer strategies across different game environments and may reduce the success rate of cooperation when different types of agents are introduced.

However, the paradigm of separating language processing from decision-making essentially contradicts the central claim of Wittgenstein's theory of language games - that there is an intrinsic unity of language, intention, and action~\cite{cai2024languageevol}.
Motivated by this, we introduce Multi-agent Kahneman \& Tversky’s Optimization (Multi-agent KTO, \textbf{MaKTO}), where the language models learn through direct interactions with different agents or models.
Specifically, MaKTO 
\textit{1)} builds on KTO for efficient training, 
\textit{2)} employs multi-agent gameplay with a \textit{diverse} model pool instead of \textit{self-play}, to enhance the generalization, and 
\textit{3)} uses \textit{stepwise} preference selection preference data selection using heuristic, voting-based, and verifier-based methods, rather than simple win-loss outcomes.
To help the model rapidly acquire domain-specific knowledge of terminology and strategies in Werewolf, our training process also incorporates behavior cloning using game-specific terms, strategy guides, and expert-annotated gameplay records with chain-of-thought~\cite{wei2022chain} before MaKTO to optimize action decision-making in gameplay.

\begin{figure*}
    \centering
    % \raggedright
    \vspace{-20pt}
    \includegraphics[width=1.01\linewidth]{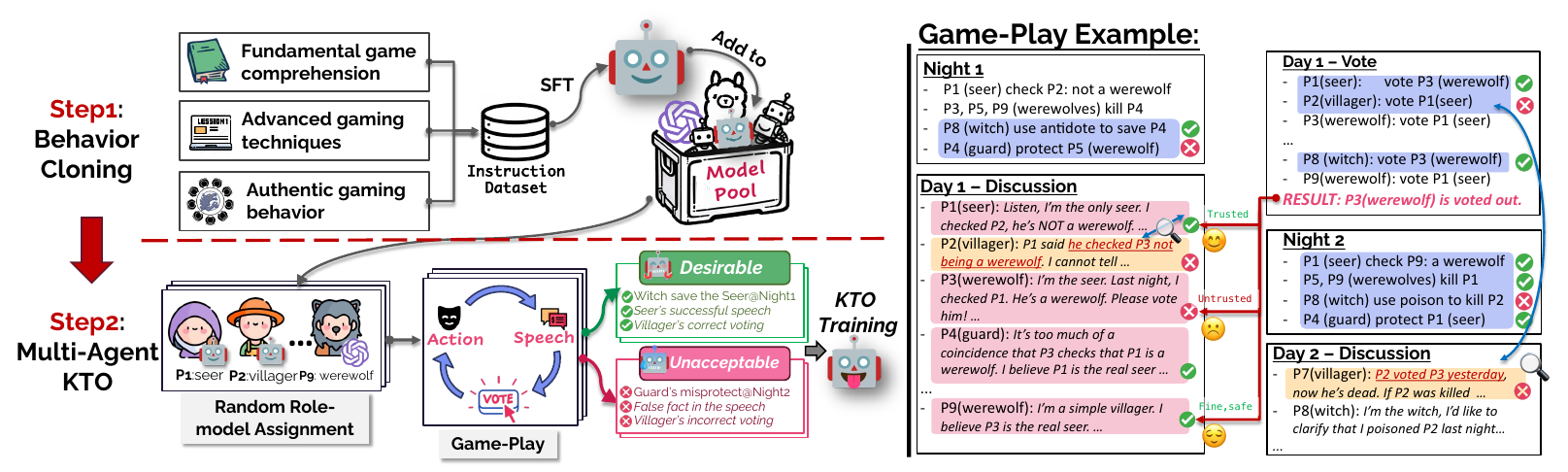}
    \caption{The overall training process consists of \textbf{(1) behavior cloning} using instruction data~($\S$\ref{sec:method_sft}) and \textbf{(2) multi-agent KTO}($\S$\ref{sec:method_kto}). In multi-agent gameplay, we randomly assign roles to agents to create diverse interactions that optimize the target model. A stepwise selection process (Right) identifies desirable and unacceptable preference data using three methods: \colorbox{myblue!50}{heuristic-based}, \colorbox{mypink!50}{staged voting-based}, and \colorbox{myyellow!50}{verifier-based} selection. The preference data is finally used for KTO optimization.}
    \label{fig:sft-kto-pipeline}
    \vspace{-18pt}
\end{figure*}

We perform various experiments, including tournament evaluation, Turing-style detectability test, behavioral analysis, generalization ability test, and ablation studies. 
The experiments show that MaKTO achieved an average win rate of 61\% in 9-player Seer-Witch-Guard games against various models such as GPT-4o, Claude-3.5, and multi-staged RL agent. This outperformed both GPT-4o and the two-stage RL agent, with relative improvements of 23.0\% and 10.9\%, respectively.
In head-to-head matches against human expert players, MaKTO reaches a comparable average win rate of 60\%.  Also, its conversational style is less distinguishable from humans, with only 48.9\% accuracy in the Turing-style detectability test.
% Additionally, we introduce a new game configuration with the role of \textit{Hunter}, which is not in the training data of MaKTO. Experiment (Sec.~\ref{sec:result:gen_to_hunter}) shows that the model performs well, demonstrating generalization ability. This generalizability suggests that our model architecture is not limited to specific \textit{Werewolf} game settings, but has the potential to be extended to a broader range of social deduction games, and even multiplayer negotiation scenarios.

Our contributions are:
\begin{itemize}[noitemsep, topsep=0pt, parsep=1pt, partopsep=0pt, leftmargin=*]
    \item We propose Multi-agent KTO~(MaKTO), a method that enhances LLMs' strategic reasoning in the game environment through \textit{multi-agent interactions}, without requiring paired data.
    
    \item We create a large-scale dataset of expert Werewolf players' utterances and actions during gameplay, as well as the abundant COT behind their decisions, allowing for effective behavior cloning and fine-tuning of LLMs.
        
    \item We perform extensive experiments to show that our model achieves human-level performance and strong generalization capabilities across different game settings.
\end{itemize}

\section{Our Approach}
\label{sec:method}
In this section, we describe in detail our training method~(Fig.~\ref{fig:sft-kto-pipeline}) in detail, including expert data collection, behavior cloning, and multi-agent KTO.

\subsection{Expert Data Collection}
\label{sec:method_data}

Despite various meticulously designed prompting methods~\cite{xu2023exploring, lan2023llm, wang2023avalon,sato2024implementation}, LLMs still exhibit a huge difference in language style and strategy play from real Werewolf players.
While Automatic Speech Recognition (ASR) systems enable the collection of textual data from online Werewolf games~\cite{wu2024enhance}, they cannot capture players' underlying reasoning.
To address these issues, we collaborate with $17$ experienced Werewolf players, including individuals with over a thousand games of experience and competitive tournament participants, to annotate the dataset.
We also ask these expert players to document their thought behind each decision during the game.
% We require advanced players to articulate their thoughts in textual format and document distinct thought processes while maintaining the authenticity of the Werewolf game.
Our dataset consists of:
% While Automatic Speech Recognition (ASR) systems enable the collection of textual data from online Werewolf games~\cite{wu2024enhance}, they cannot capture players' underlying reasoning.
% the informal nature of spoken communication and the absence of explicit reasoning processes underlying decision-making remain as challenges. 
% Therefore, we require advanced players to articulate their thoughts in textual format and document distinct thought processes while maintaining standardization and authenticity of the Werewolf game. Specifically, our annotated dataset encompasses \textbf{game regular record} and \textbf{thinking process data}: 

\noindent\textbf{Gameplay record}: Contains the nighttime action records of special-role villagers and werewolves, daytime speeches and votes of players, and post-game reviews.

\noindent\textbf{Thinking process annotation}: Documents players' reasoning for:
% action, speech, and voting constitute the core elements of the Werewolf game. We instruct expert players to annotate distinct thinking processes for each of these three components.
\begin{itemize}[noitemsep, topsep=0pt, parsep=1pt, partopsep=0pt, leftmargin=*]
    \item \textbf{Action}: the rationale behind night actions (e.g., seer's checks, werewolf's kills)
    % this section primarily annotates the rationale behind night actions, such as the seer's intention behind checking a certain player, or the werewolf's motive for killing.
    \item \textbf{Speech}: outline of the daytime speech, the identity assessments for other players, and the call for the vote.
    % this section primarily summarizes the content expressed after the statements, including the identity accorded to other players during the daytime speech and the call for the vote. It can be regarded as an outline of the daytime speech.
    \item \textbf{Voting}: detailed reasons for voting and player identity predictions; Players are also required to distill the day's events into a consolidated record, create notes, and formulate a rudimentary strategy for the next game phase.
\end{itemize}

We collected $331$ annotated \textit{Werewolf} games from $17$ expert players via our platform, with 51 additional games reserved for LLM evaluation in Sec.~\ref{sec:result:behavior}. Detailed statistics are in Appendix~\ref{sec:app:data_stats}.
% 3. stats: how many samples, tokens,
% ## summary sentence for overall + table: list detail numbers 
% 17 expert players provide annotations using our self-built annotation platform. 
% We collect 331 Werewolf games for training, including 278 9-player games (Seer, Witch, Guard or Hunter) and 53 7-player games (Seer, Guard or Witch).
% We collect 331 matches of Werewolf games annotated by 17 advanced players based on our self-built annotation platform. 
% This dataset includes 278 9-player games (using the setups of Seer, Witch, Guard and Hunter) and 53 7-player games (using the setups of Seer, Guard and Witch). 
% The specific rules for different game settings are provided in Appendix~\ref{app:game_rules}. Each game features randomly involved participants and randomly assigned Werewolf roles to guarantee data diversity. The total duration of the game annotated by the players exceeds 1,000 hours, including 3,759 speech data entries (exceeding 540,000 tokens), 2,698 action events, and 3,875 voting records.
% \input{Tables/game_stastic}
% We also leave 51 games with voting records and role predictions for offline evaluating the basic ability of LLM in Section~\ref{sec:result:behavior}.

\subsection{Behavior Cloning}
\label{sec:method_sft}

Due to the scarcity of high-quality data in \textit{Werewolf} domain, existing LLMs generally lack a profound understanding and do not possess sufficient reasoning logic to support advanced gameplay. 
We address this by creating a comprehensive, multi-level instruction dataset for supervised fine-tuning (SFT), as shown in Fig.~\ref{fig:sft-kto-pipeline}-Step 1.
% To address this limitation, we construct a comprehensive, multi-level instruction dataset to supervised fine-tune~(SFT) the model, as illustrated in Fig.~\ref{fig:sft-kto-pipeline}(a). This approach enables LLMs to learn specific patterns and nuances by adapting their parameters according to the Werewolf data distribution.
Our instruction data is derived from three sources:
% These diverse sources contribute to the LLM's multifaceted comprehension of gaming from various perspectives.

\noindent\textbf{1. Fundamental game comprehension}: Primarily focuses on the explanation of game terminology and jargon. As players continuously innovate within \textit{Werewolf}, specialized shorthand terms for efficient communication have emerged. For example, The term ``Goldwater" refers to players verified as innocent through Seer's investigation.

\noindent\textbf{2. Advanced gaming techniques}: Collected from experienced players' online strategies, providing guidance for common game scenarios. For instance, it includes expert tips on how werewolves can effectively impersonate the Seer role and mislead the villagers.
% is obtained from online amassed strategies shared by experienced players. This content provides targeted guidance for various common scenarios encountered in the game. For instance, it offers advice on how werewolves can effectively impersonate the Seer role and mislead the villagers.

\noindent\textbf{3. Authentic gaming behavior}: Derived from expert-annotated gameplay data in Section~\ref{sec:method_data}. Benefiting from the annotated thinking process, we structure them into a ``think-before-respond" format, enabling the model to truly comprehend the logic behind each stage of the game. For action, we first output the reason and then the target object. For speech, we output expected labels for others and voting intentions as the outline before generation. For voting, we output a summary of the day's events, followed by the chain-of-thought and the voting target. See Appendix~\ref{sec:app:prompt} for the examples of the data format. Additionally, we designed a role prediction auxiliary task, which involves predicting each player's role at the end of each day based on known speech, voting, and elimination information.

% Based on expert-annotated gameplay data from Section~\ref{sec:method_data}. Using \textbf{Thinking process data}, we implement a "think-before-respond" format for actions, speech, and voting. We also include a role prediction task based on daily game information.

% is derived from high-quality real-player combat data annotated by expert players in Section~\ref{sec:method_data}. Benefiting from the annotated \textbf{Thinking process data}, we structure this practical data into a "think-before-respond" format, enabling the model to truly comprehend the logic behind each stage of the game --- action, speech, and vote. For action, we require the output of the rationale first, followed by the target object. For speech, we demand the output of intended labels for others and voting intentions before generating. For vote, we necessitate a summary of the day's events, followed by the voting rationale and the voting target.  
% Additionally, we designed a role prediction auxiliary task, which involves predicting each player's role at the end of each day based on known speech, voting, and elimination information.

\subsection{Multi-agent KTO}
\label{sec:method_kto}
Although the SFT training enhances the model's comprehensive understanding of the Werewolf game, it faces two major challenges: \textit{the characteristics of the game} and \textit{limitations in human-annotated data}.
In Werewolf, individual actions (like Seer's claims or Witch's poisoning) subtly influence outcomes, yet team victory doesn't ensure optimal individual play. 
% In Werewolf, the impact of a single player's actions on the outcome is often subtle, and a faction's success doesn't guarantee that every member made smart choices. 
% For example, the Seer's persuasive speech or the Witch's critical poisoning sway the situation.
Additionally, even expert data includes both good and poor decisions from winners and losers, making it challenging to evaluate individual choices based solely on game outcomes.
% annotated data from experts inevitably contains decisions of varying quality, because it includes both winners and losers. These factors collectively make it difficult to accurately assess the quality of individual decisions based on win-loss outcomes, which in turn affects the model's performance.
To address these challenges, we develop \textbf{Multi-agent KTO (MaKTO)} to mitigate such suboptimality. 
MaKTO features three key aspects: 
1) It employs the Kahneman-Tversky Optimization (KTO) algorithm~\cite{ethayarajh2024kto} for decision refinement.
2) It adopts multi-agent gameplay to get diverse training data.
% 2) It adopts multi-agent gameplay, where the reference model participates in gameplay alongside various other models to obtain diverse training data.
3) Instead of optimizing the \textit{entire} trajectory based on win/loss outcomes, it optimizes step-wise policies.

\paragraph{Kahneman-Tversky Optimization}
We argue that KTO is particularly suitable for such a multi-agent language game for two reasons.
1) Multi-agent environments are more complex than single/two-agent scenarios, where single-agent interactions with the environment often yield clear feedback, while interactions between agents can have countless possibilities. Moreover, multi-agent dialogues have huge action spaces, leading to sparse trajectory sampling. This makes online reinforcement learning algorithms slow to converge and hard to train. KTO, similar to offline RL, offers a viable solution.
2) Unlike preference optimization algorithms such as DPO~\cite{rafailov2024direct} and its variants~\cite{hong2024orpo,pal2024smaug,meng2024simpo,lai2024step,song2024trial}, it is nontrivial to get ``\texttt{prompt-chosen-reject}'' paired data. However, you can determine whether an output is acceptable or not through game rules and feedback from other agents. KTO, not requiring paired data for training, makes it ideal for such \textit{``try once''} scenario in multi-agent preference optimization.
The loss function of KTO is in Appendix~\ref{sec:app:makto_details}.
% We assign desirable and undesirable behaviors manually and use KTO to update the model. 
% Given the prompt-response from dataset \((x, y) \in D\), KTO uses the following loss to optimize the policy \(\pi_\theta\):

% \[r_\theta(x,y) = \log \frac{\pi_\theta(y|x)}{\pi_{ref}(y|x)} \]

% \[z_0 = \mathbb{E}_{(x,y)\sim D}[KL(\pi_\theta(y|x) \| \pi_{ref}(y|x)) ]  \] 

% \[
% v(x,y) = 
% \left\{
% \begin{array}{ll}
% \lambda_D\sigma(\beta(r_\theta(x,y)-z_0)), & \text{if } y \sim y_{desirable}|x \\
% \lambda_U\sigma(\beta(z_0-r_\theta(x,y))), & \text{if } y \sim y_{undesirable}|x \\
% \end{array}
% \right.
% \]

% Therefore, 

% \[L(\pi_\theta;D) = \mathbb{E}_{(x,y)\sim D}[\lambda_y - v(x,y)]\] 

% Here, \( \lambda_D \) and \( \lambda_U \) are hyperparameters for the desirable and undesirable losses, respectively. 
% The parameter \( \lambda_y \) represents \( \lambda_D \) when \( y \) is desirable and \( \lambda_U \) when \( y \) is undesirable.

\paragraph{Multi-agent Gameplay}~
We find that in multi-agent settings, merely SFT or \textit{self-play} can lead to rigid strategies and poor generalization~(analysis in Sec.~\ref{sec:result:ablation}).
For example, we find that although the model performs well in scenarios with \textit{two} players claiming to be the seer, its performance significantly deteriorates when only \textit{one or more than three} players claim this role. 
% This lack of robustness is inadequate for the game's requirements.
We argue that in multi-agent environments, the diversity of peers and opponents is crucial. Therefore, instead of relying on self-play, we employ a multi-agent play using a diverse model pool --- including various SFT models (\texttt{Llama3.1}, \texttt{Qwen2.5}), off-the-shelf LLMs (\texttt{GPT-3.5}, \texttt{GPT-4o}, and \texttt{Claude}), as well as the agent that uses RL for decision-making and LLM for speech generation.
% As shown in Fig.~\ref{fig:sft-kto-pipeline}(b), we create a diverse model pool comprising SFT models based on various base models like \texttt{Llama3.1} and \texttt{Qwen2.5} with different model sizes, a range of off-the-shelf LLMs (such as \texttt{GPT-3.5} and \texttt{GPT-4o}, \texttt{Claude} etc.), as well as the agent that uses RL for decision-making and LLM for speech generation.
We then randomly select models from the model pool and assign them to different roles in the game. The multi-agent interactions allow for the exploration of a broader strategy space and help avoid overfitting to specific policies or patterns.
% , thus enhancing the model's robustness.

\paragraph{Stepwise Preference Data Selection}~
Defining all actions in an agent's trajectory as desirable or not based solely on the faction's win/loss result is too simplistic.
Fortunately, Werewolf's alternating day-night gameplay allows for a more nuanced selection of the desirable and unacceptable step-wise process policies.
% As shown in Fig.~\ref{fig:sft-kto-pipeline}(b) and Table~\ref{tab:kto_selection}, 
Specifically, we employ three methods:

\begin{itemize}[noitemsep, topsep=0pt, parsep=1pt, partopsep=0pt, leftmargin=*]
    \item \textbf{Heuristic-based selection} 
    identifies actions based on role-specific strategies and game rules, focusing primarily on nighttime actions and voting phases. For example, werewolves targeting special roles is desirable while not attacking is unacceptable, and for witches, successfully poisoning werewolves is desirable. For voting, unified voting of the villagers against werewolves is preferred, while infighting or vote-splitting, thus weakening the villager team's position, is discouraged.
    
    \item \textbf{Staged voting-based selection} uses voting outcomes to assess speech quality. Intuitively, players who voted out likely gave suboptimal speeches, either failing to defend themselves or contradicting others' observations. Special role players (seer, witch) face stricter evaluation, with receiving majority villager votes considered unacceptable.
          
    \item \textbf{Verifier-based selection} employs strong external LLMs, like GPT-4o to verify speech consistency with game facts and events. This can reduce the hallucination by performing fact consistency checks, as well as logical coherence and self-contradictory.
    
    % evaluates whether the model-generated speeches align with known facts in the game by introducing additional strong off-the-shelf LLMs, such as Claude or GPT-4o. This approach addresses hallucination issues in our SFT model. The verifier performs fact consistency checks, \textit{e.g.}, whether a player's speech contradicts their role and known game events. It also validates logical coherence to detect whether the generated content contains self-contradictory statements or not.
    
\end{itemize}
For an explanation of the selection criteria, especially Heuristic-based selection, you may refer to Table~\ref{tab:kto_selection} in Appendix~\ref{sec:app:makto_details}.

% \input{Tables/kto_data}

% \todo{put training details in appendix}

% \subsection{Training Details}

% % \bmhead{Base model}~
% Given that our collected expert training data is based on Chinese, and considering a stronger understanding of the Chinese context, we choose \texttt{Qwen2.5-14b-instruct} and \texttt{Qwen2.5-72b-instruct}~\cite{yang2024qwen2} as the base models for training. 
% % \bmhead{SFT}~
% The SFT dataset comprises 25k samples, including 380 samples of fundamental game comprehension data with terminology explanations, 372 Q\&As on advanced gaming techniques, 12k annotated authentic gaming behavior data, and 12k general SFT corpus. 
% We employed DeepSpeed ZeRO-3 optimization with a learning rate of 1e-6, a warm-up ratio of 0.05, and trained for 3 epochs.
% % \bmhead{MaKTO}~
% For the MaKTO phase, we collected 20k preference data entries from the Seer-Witch-Guard games, consisting of 12k desirable and 8k unacceptable samples. The model pool includes \texttt{GPT-4o\_mini}, \texttt{GPT-4o}, fine-tuned \texttt{Qwen2.5-14b-instruct}, fine-tuned \texttt{Llama-3.1-8B-Instruct}, and fine-tuned \texttt{Qwen2.5-72b-instruct}. 
% We set the KTO hyper-parameters with  \( \lambda_D =0.7\) and \( \lambda_U=1.0 \). 
% The training utilized DeepSpeed ZeRO-3 optimization, with a learning rate of 1e-6, a batch size of 2 per device, 150 warmup steps, and train for 20 epochs.
% We utilized 8 A100 GPUs for the 14B model and 32 A100 GPUs for the 72B model.

\section{Experiments}
\label{sec:experiment}
% or use "sidewaystable"
\begin{table*}[t]
    \centering
    \small
    \begin{tabular}{l|rrr|r|rrr|r}
    \toprule
      &
    \textbf{GPT4o\_mini} & \textbf{GPT4o} &  \textbf{Claude} & 
    \textbf{Mix} & \textbf{SFT-14b} & \textbf{SFT-72b} & \textbf{MaKTO-72b} & \textbf{Avg.} \\ 
    \midrule
    \textbf{GPT4o\_mini} & {\color[HTML]{FF0000} 0.50} & 0.44 & 0.23 & 0.13 & 0.23 & 0.24 & 0.12 & 0.270 \\
    \textbf{GPT4o} & \textbf{0.56} & {\color[HTML]{FF0000} 0.50} & 0.66 & \textbf{\underline{0.56}} & 0.44 & 0.40 & 0.35 & 0.496 \\
    \textbf{Claude} & \textbf{0.77} & 0.34 & {\color[HTML]{FF0000} 0.50} & 0.46 & 0.48 & 0.44 & 0.38 & 0.481 \\ 
    \midrule
    \textbf{Mix}~\cite{wu2024enhance} & \textbf{0.87} & 0.44 & \textbf{0.54} & {\color[HTML]{FF0000} 0.50} & \textbf{\underline{0.58}} & 0.45 & 0.47 & 0.550 \\ 
    \midrule
    % qwen14b & very bad \\
    % qwen72b \\
    \textbf{SFT-14b} & \textbf{0.77} & \textbf{0.56} & \textbf{0.52} & 0.42 & {\color[HTML]{FF0000} 0.50} & \textbf{0.57} & 0.49 & 0.547 \\
    \textbf{SFT-72b} & \textbf{0.76} & \textbf{0.60} & \textbf{0.56} & \textbf{0.55} & 0.43 & {\color[HTML]{FF0000} 0.50} & 0.42 & 0.546 \\
    \textbf{MaKTO-72b} & \textbf{\underline{0.88}} & \textbf{\underline{0.65}} & \textbf{\underline{0.62}} & \textbf{0.53} & 0.51 & \textbf{\underline{0.58}} & {\color[HTML]{FF0000} 0.50} & \textbf{0.610} \\ 
    \bottomrule
    \end{tabular}
    \caption{Average win rates of the models in the Seer-Witch-Guard setting. A win rate above 0.5 (in \textbf{bold}) indicates that the model in the row significantly outperforms the model in the column.}
    \label{tab:avg_win_9guard}
    \vspace{-15pt}
\end{table*}

\label{sec:result}

In this section, we first evaluate our approach in 9-player Seer-Witch-Guard Werewolf games through tournaments, human-AI competitions, and Turing-style detectability tests. 
We then analyze MaKTO's superior performance through behavioral studies, test its generalization capability with a new \textit{Hunter} role in a new game setting, 
and validate key components through ablation studies.
The implementation details are in Appendix~\ref{sec:app:train}.

\subsection{Tournament Win Rate Evaluation}
\label{sec:result:winrate}

\subsubsection{Inter-agent Tournament}
\label{sec:result:inter_agent_win}
\noindent\textbf{Experimental Setup}
We evaluate our approach against several strong baselines: API-based LLM agents (\texttt{GPT-4o}, \texttt{GPT-4o-mini}, and \texttt{Claude-3.5-Sonnet}\footnote{Model Versions: \textbf{GPT4o-mini}: gpt-4o-mini-2024-07-18, \textbf{GPT4o}: gpt-4o-2024-08-06, \textbf{Claude-3.5-Sonnet}: claude-3-5-sonnet-20241022.} using chain-of-thought prompting~\cite{wei2022chain}), the \texttt{Mix} agent combining LLM with RL policy for decision making following~\cite{wu2024enhance}, and SFT models (based on \texttt{Qwen2.5-14b/72b-instruct}) trained on our expert-annotated dataset. 
% Our experiments include both head-to-head competition and random competition including randomly assigned roles from the models described above.

% \input{Tables/avg_win_9hunter}

\begin{figure}
    \centering
    \includegraphics[width=0.9\linewidth]{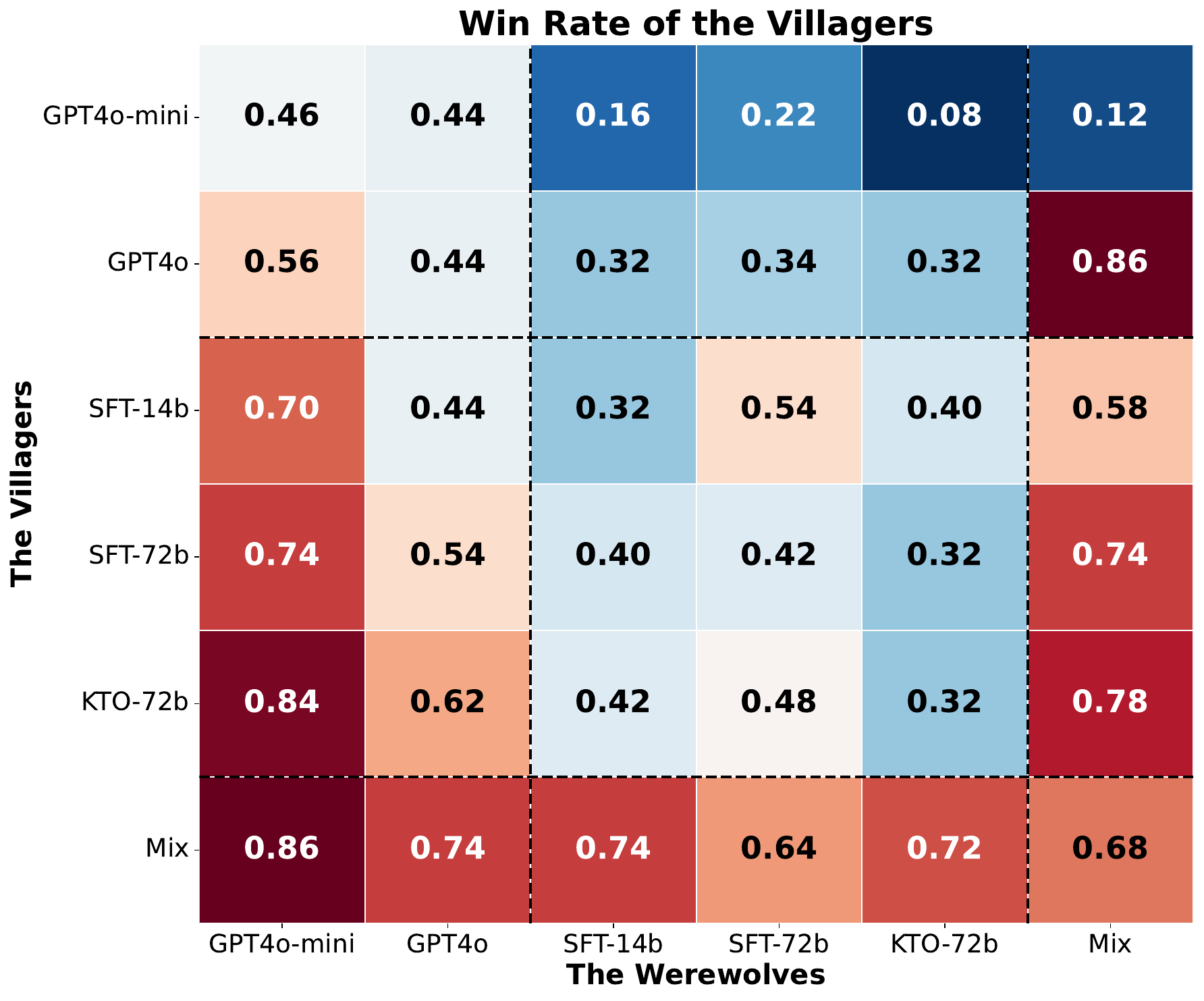}
    \caption{Villager win rate matrix of the head-to-head competition: villager models (y-axis) vs. werewolf models (x-axis). \textbf{Lower left}: higher values show stronger villager performance; \textbf{Upper right}: \underline{lower} values indicate stronger werewolf performance.}
    \label{fig:villager-win-matrix}
\end{figure}

\noindent\textbf{Head-to-head Competition}
In head-to-head competitions, where one model controls the entire villager team (6 agents) and another controls the werewolf team (3 agents), MaKTO-72b achieved a 61\% average win rate across 100 games (50 games per faction), significantly outperforming all baselines (Table~\ref{tab:avg_win_9guard}). Notably, while the Mix agent showed strong performance as villagers, it struggled as werewolves due to overly aggressive strategies and policy contradictions (Fig.~\ref{fig:villager-win-matrix}). 
When comparing SFT models of different sizes (14B and 72B), we observed similar win rates but significantly fewer factual hallucinations in the 72B model's generated speeches.
% indicating that win rates alone don't fully reflect model performance.

\begin{figure}
    \centering
   \begin{minipage}{1\linewidth}
        \centering
        \subfloat[TrueSkill Score of models]
        {
          \centering
          \label{fig:rc_trueskill}
          \includegraphics[width=0.55\textwidth]{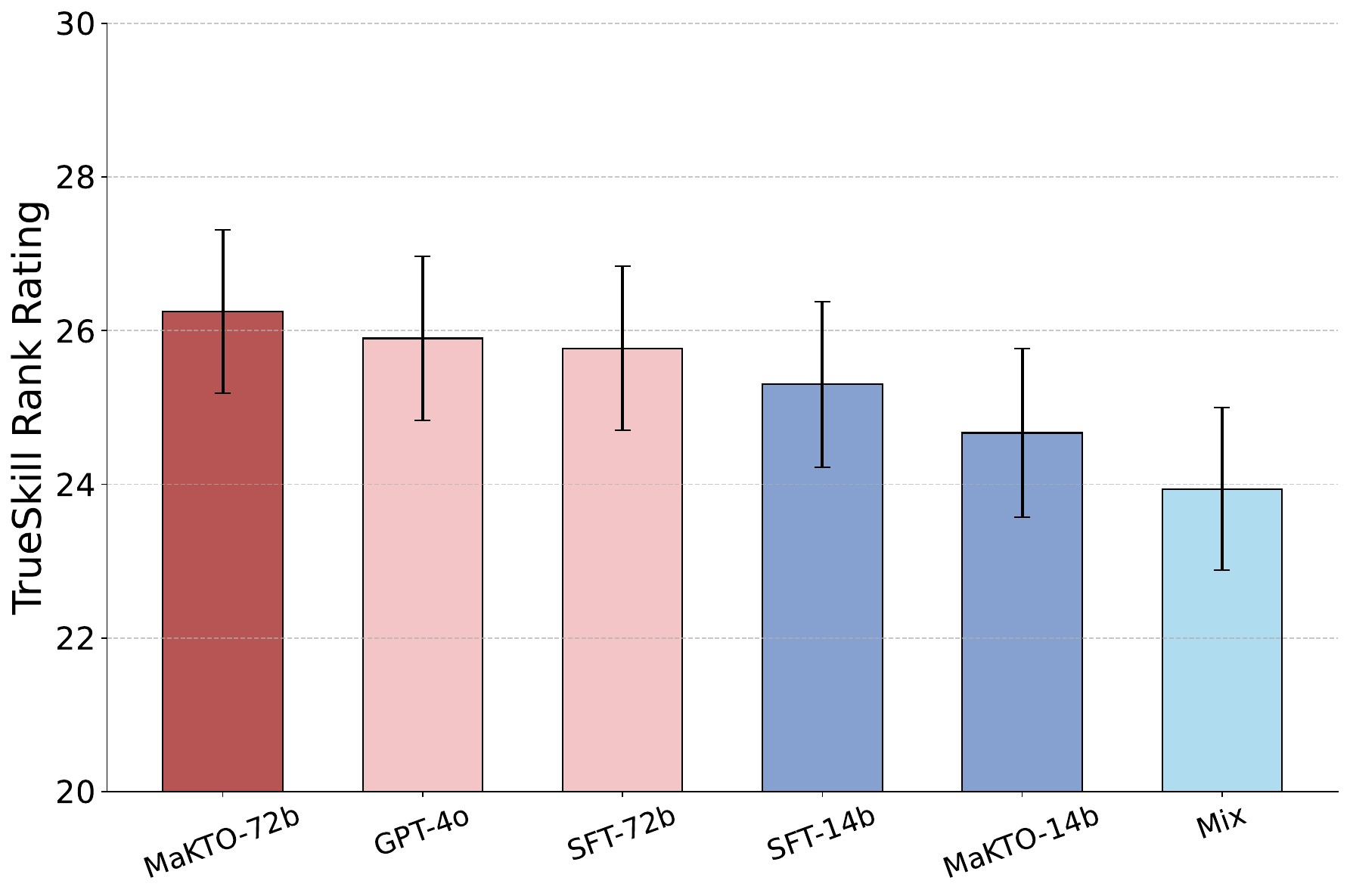}
        }
        \subfloat[Role Winning rates]
        {
          \centering
          \label{fig:rc_winningrate}
          \includegraphics[width=0.38\textwidth]{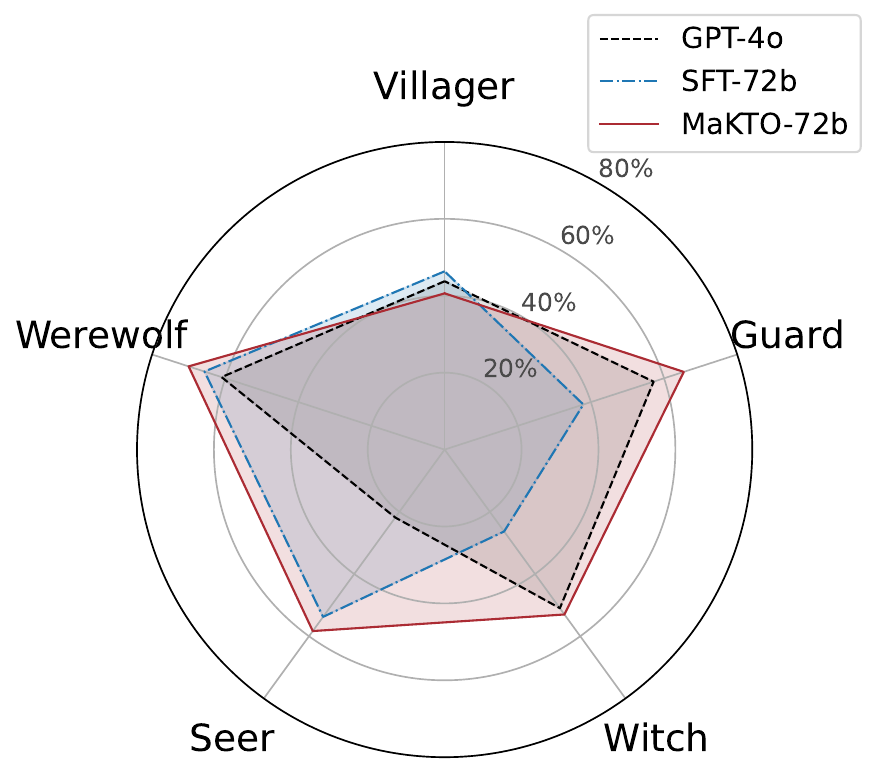}
        }
    \end{minipage} 
    \caption{Results of 260 random competitions in 9-player Seer-Witch-Guard game setting.}
    \label{fig:random_competition}
\end{figure}

\noindent\textbf{Random Competition}
In random competitions with diverse role assignments across 260 games, MaKTO-72b achieved the highest TrueSkill rating~\cite{herbrich2006trueskill} (Fig.~\ref{fig:random_competition}). This format better reflects the model's adaptability across different roles and team compositions. Particularly, MaKTO-72b significantly outperformed \texttt{GPT-4o} when playing as the Seer, suggesting more persuasive statements and better trust-building capabilities. Compared to SFT-72b, MaKTO-72b also exhibited higher winning rates when playing as Guard and Witch, demonstrating better strategic skill usage.

%%%%%%%%%%%%%%%%%%%%%%%%%%%%%%%%%%%%%%%%%%%
\subsubsection{Human-AI Tournament}
\label{sec:result:human_agent_win}

\noindent\textbf{Experimental Setup}
To evaluate our model's performance in real-world games, we conduct human-AI tests with 14 experienced human players (1000+ games each) through head-to-head and random competition.
% 14 experienced human players (each with over 1,000 games of experience), assessing both performance and human-likeness through three complementary evaluations. 
Unlike previous studies~\cite{xu2023language,wu2024enhance} that only introduced single AI or human players, we involve multiple human players and AI models, creating more challenging and realistic environments.
% The human evaluation also includes head-to-head competitions and random competitions.

% \begin{figure}
%     \centering
%     \includegraphics[width=0.5\linewidth]{Figures/human_random_all.pdf}
%     \caption{Win rate of players in random competition. H1-H14 stand for the win rate of human players.}
%     \label{fig:human_eval_random}
% \end{figure}

\noindent\textbf{Head-to-head Competition}~
We first evaluate MaKTO in head-to-head competition for 20 Seer-Witch-Guard games (10 games for each side), where MaKTO-72b plays all villagers or werewolves independently, and the opposing side are played by humans. 
% As the evaluation goes, we find that the textual style between bots and humans is different. The difference will cause the game to become unbalanced, as players can distinguish who werewolves are according to textual style. Therefore, we add two bots to the villager team when bots playing werewolves and vice versa. 
MaKTO-72b achieved a 60\% win rate (5/10 as villagers, 7/10 as werewolves; 
Table~\ref{tab:human_head_to_head_eval}), 
indicating that it is on par with high-level human players.
\begin{table}[h]
    \centering
    \small
    \begin{tabular}{ll|cc}
    \toprule
    \multicolumn{2}{l|}{\multirow{2}{*}{}} & \multicolumn{2}{c}{Humans} \\ 
    \cmidrule(lr){3-4}
    \multicolumn{2}{l|}{} & \multicolumn{1}{l|}{Villager} & \multicolumn{1}{l}{Werewolf} \\ 
    \midrule
    \multicolumn{1}{c|}{\multirow{2}{*}{\textbf{MaKTO-72b}}} & Villager & \multicolumn{1}{c|}{--}       & 0.5 \\ 
    \cmidrule(lr){2-4}
    \multicolumn{1}{c|}{} & Werewolf & \multicolumn{1}{c|}{0.7} & --                           \\ 
    \bottomrule
    \end{tabular}
    \caption{Win rate of MaKTO-72b in head-to-head competition with humans in 9-player Seer-Witch-Guard setting.}
    \label{tab:human_head_to_head_eval}
\end{table}

\begin{figure}
    \centering
    \includegraphics[width=\linewidth]{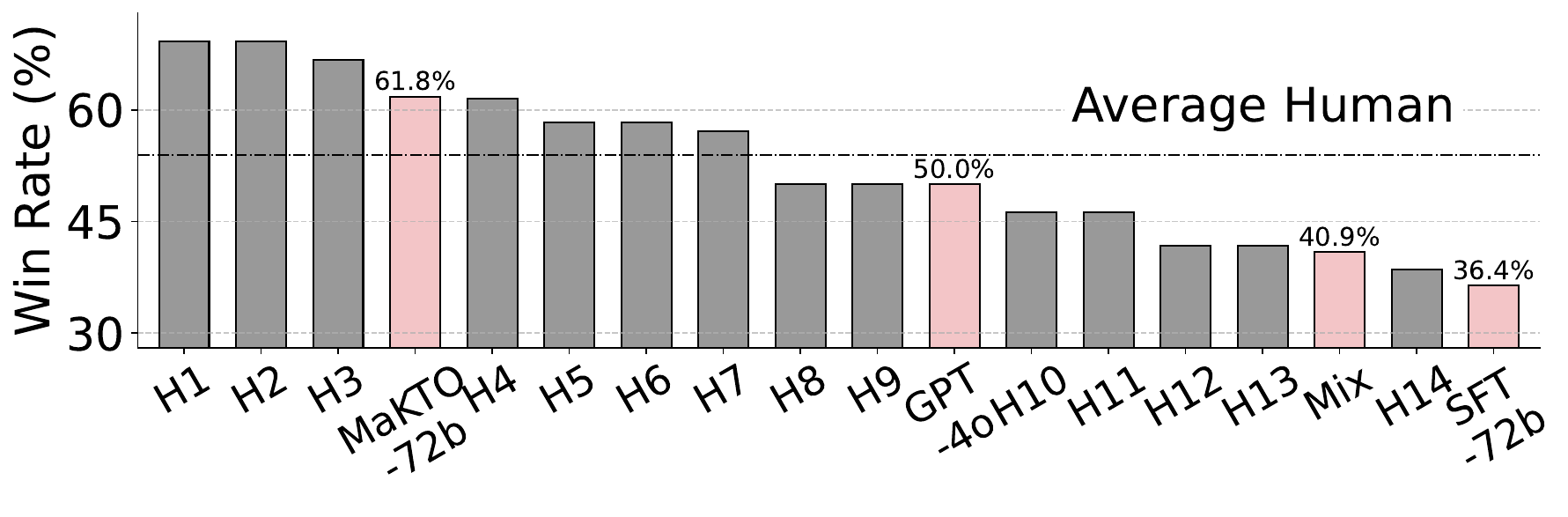}
    % \vspace{-25pt}
    \caption{Win rate of players in random competition. H1-14 stand for the win rate of human players.}
    \label{fig:human_eval_random}
    % \vspace{-15pt}
\end{figure}

\noindent\textbf{Random Competition}~
We also evaluate in random competition for 30 seer-witch-guard 9-player games, where each player is randomly selected from 2-7 human players and the model pool (including \texttt{ GPT-4o}, Mix agent, SFT-72b, and MaKTO-72b). 
% Our setup involves 14 experienced human players (each with over 1,000 games played) and various AI models from a diverse model pool (including \texttt{ GPT-4o}, Mix agent, SFT-72b, and MaKTO-72b). 
% In each 9-player game, 2-7 players were randomly assigned to AI models selected from this pool, creating diverse human-AI mixed scenarios that better reflect real-world interactions and test models' adaptability against both human players and different AI agents.
As in Fig.~\ref{fig:human_eval_random}, MaKTO-72b achieved a win rate 61.8\%{\footnotesize$\pm$8.3\%} win rate, ranking fourth among all players and surpassing the average human win rate of 54\%. This shows that MaKTO-72b has strong adaptability against both human and AI opponents.

% has not only achieved human-level competence in Werewolf, but also exhibits robust adaptability to diverse game situations, strategies, and playing styles from both human and AI opponents.

% \begin{figure*}[htbp]
%     \centering
%     \begin{minipage}[b]{0.7\linewidth}
%         \centering
%         \includegraphics[width=\linewidth]{Figures/human_random_all.pdf}
%         \caption{Win rate of players in random competition. H1-14 stand for the win rate of human players.}
%         \label{fig:human_eval_random}
%     \end{minipage}
%     \hfill
%     \begin{minipage}[b]{0.27\linewidth}
%         \centering
%         \includegraphics[width=\linewidth]{Figures/human_detect.pdf}
%         \caption{Model's human recognition accuracy. 
%         % Lower rate indicates that the model is more human-like and indistinguishable.
%         }
%         \label{fig:human_dect}
%     \end{minipage}
% \end{figure*}

\subsection{Turing-style Detectability Test}
\label{sec:result:turing}

% \begin{figure}
%     \centering
%     \includegraphics[width=0.5\linewidth]{Figures/human_detect.pdf}
%     \caption{Model's human detection rate. A lower rate indicates that the model is more human-like and indistinguishable.}
%     \label{fig:human_dect}
% \end{figure}

% begin{minipage}[b]{0.27\textwidth}
%         \centering
%         \includegraphics[width=\linewidth]{Figures/human_detect.pdf}
%         \caption{Model's human detection rate. 
%         % Lower rate indicates that the model is more human-like and indistinguishable.
%         }
%         \label{fig:human_dect}
%     \end{minipage}
% \end{figure}

% \begin{figure}
%     \centering
%     \includegraphics[width=0.7\linewidth]{Figures/human_detect.pdf}
%     \caption{Model's human detection rate. Lower rate indicates that the model is more human-like and indistinguishable.}
%     \label{fig:human_dect}
% \end{figure}

We conduct rigorous Turing-style blind detectability tests in both competitions.
We require each human player to explicitly judge whether \textit{every other} participant is human or AI, without any prior knowledge of AI presence. 
This mandatory assessment provides a rigorous human similarity assessment.
MaKTO achieves detection accuracy of only 48.9\% (Tab.~\ref{tab:human_dect}), lower than random chance, indicating that our model successfully passes this specialized Turing test by convincingly emulating human-like gameplay characteristics and social behaviors.
On the contrary, \texttt{GPT-4o} has a much higher detection rate (76.6\%) due to the significant differences in speaking style and voting behavior from human players.

\begin{table}[htb]
    \centering
    \small
    \resizebox{\linewidth}{!}{
    \begin{tabular}{c|rrrr}
        \toprule
         & \textbf{GPT-4o} & \textbf{Mix} & \textbf{SFT-72b} & \textbf{MaKTO-72b} \\
         \midrule
        \textbf{Detection Acc.} & 76.6\% & 76.1\% & 53.0\% & \textbf{48.9\%} \\
        \bottomrule
    \end{tabular}
    }
    \caption{Model's human detection accuracy. Lower rate indicates that the model is more human-like and indistinguishable.}
    \label{tab:human_dect}
\end{table}

\subsection{Behavioral Analysis}
\label{sec:result:behavior}

\subsubsection{Comparison with Baseline Model}
\label{sec:result:behavior_model}

% As shown in Figure~\ref{fig:villager-win-matrix}, MaKTO outperformed SFT models in both werewolf and villager win rates. 
In order to understand why MaKTO has a higher win rate in tournaments, we analyze the behaviors and decisions generated during the tournament. 
For a fair comparison, we selected \texttt{GPT-4o} as the opponent. We run 50 games between both sides and compute the proportion of behavior occurrences as metrics.
Fig.~\ref{fig:behavior_bot} shows the results of the metrics. Detailed data and explanations of the metrics are in Appendix~\ref{sec:app:behavior}.
% with detailed statistical data and the explanation of the metrics available in Appendix \todo{appendix}.

% To understand KTO's performance in detail, we examined its decision-making at each stage. 
% For a fair comparison, we choose \texttt{GPT-4o} as the opponent in the behavioral analysis, and run 50 games for each side.

\paragraph{When the trained models played as villagers against \texttt{GPT-4o} werewolves,}
we evaluate voting accuracy (\textit{Vote Acc.}).
% and abstention rate (\textit{Abstention}). 
For special roles, we examine: Seer's werewolf identification accuracy (\textit{Werewolf Check}), Witch's first-night rescue rate (\textit{Save @ Night 1}), werewolf poisoning accuracy (\textit{Correct Poison}), 
% villager mispoison rate (\textit{Mispoison}), 
and Guard's special role protection rate (\textit{Protect God}).
% and werewolf protection errors (\textit{Misprotect}).
Fig.~\ref{fig:behavior_bot} (left) shows that MaKTO outperforms the baseline SFT model across all the metrics, which yields a higher villager win rate. 
These improvements can be attributed to the stepwise decision rewards and penalties in the MaKTO training.

\begin{figure}
    \centering
    \includegraphics[width=\linewidth]{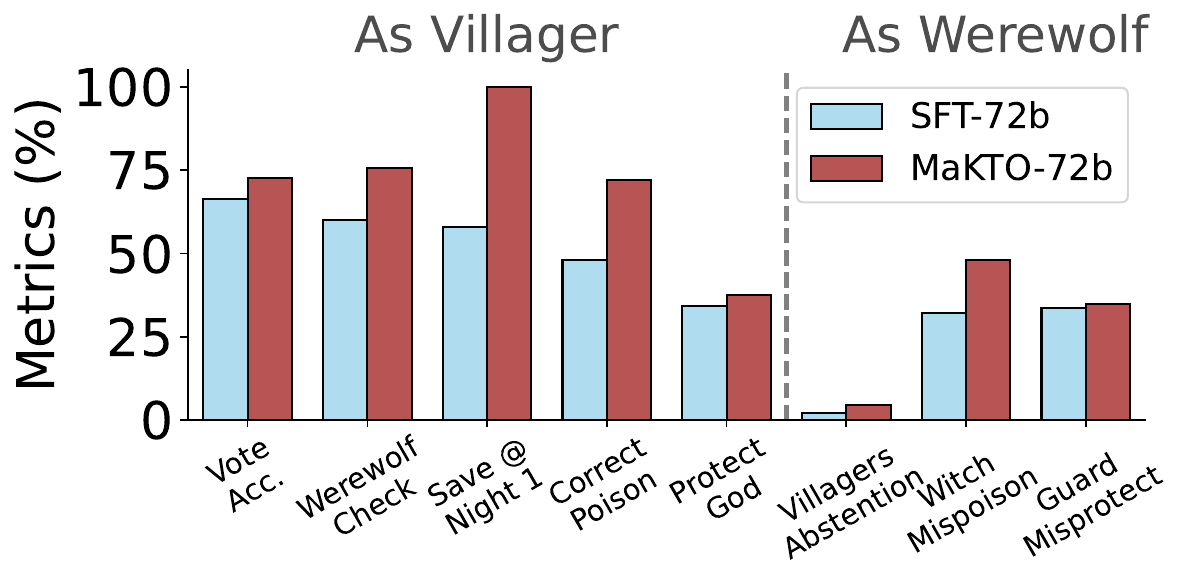}
    % \vspace{-25pt}
    \caption{Behavioral Analysis: model performance as villagers (left) and werewolves (right) against GPT-4o. For werewolf, \underline{\textbf{lower} opponent} scores indicate better performance of the model. See Appendix~\ref{sec:app:behavior} for exact values.}
    \label{fig:behavior_bot}
    % \vspace{-15pt}
\end{figure}

\paragraph{When our models played as werewolves against GPT-4o villagers}, we evaluated the opponent's behavior instead of our model's direct performance. The rationale behind this approach is that better werewolf deception leads to more opponents' confusion and mistakes. Specifically, we measured the following metrics of the \texttt{GPT-4o} opponents:
\textit{1)} Voting abstention rate~(\textit{Abstention}): Higher rates indicate difficulty identifying werewolves.
\textit{2)} Seer's werewolf identification rate~(\textit{Seer Check}, see Table~\ref{tab:kto_fine_grain_wolf} for specific values): Lower rates suggest successful misdirection.
\textit{3)} Witch's villager poisoning rate~(\textit{Mispoison}): Higher rates indicate better werewolf concealment.
\textit{4)} Guard's werewolf protection rate~(\textit{Misprotect}): Higher rates suggest effective deception.
All the fine-grained metrics of MaKTO are better than the SFT model, which shows that MaKTO werewolf has a superior camouflage capability in the game.

\subsubsection{Comparison with Human}
A feature of the \textit{Werewolf} game lies in its prevalence of deception, particularly among the werewolf players. Werewolves never openly admit their identity in daytime discussions; instead, they make up various identities and stories to protect themselves. 
It becomes important for villagers to correctly predict who is the werewolf, so that they can cast the right vote. So in this experiment, we compare the correct judgments of the villagers and compare them with humans.

We separate 51 matches from the annotated data as the test set, excluding them from the training dataset. The evaluation covers 484 voting events and 5130 identity predictions. 
The results are presented in Table~\ref{tab:static_test}.
In terms of voting, we evaluate the voting accuracy, that is, the accuracy of gods and villagers voting for werewolves; and the abstention rate. MaKTO-72b achieves the highest voting accuracy. 
In terms of identity prediction, we evaluate the accuracy of side alignment (\textit{Align. Acc.}), that is, correctly predicting gods and villagers as the good identity and werewolves as the bad identity; and the F1-score in predicting werewolves~(\textit{Wolf-pref. F1}). 
The trained models show significant improvement over the base model like \texttt{Qwen2.5-14b} and \texttt{-72b-instruct}, which demonstrate the effectiveness of the expert data we collected. Also, we notice that, the SFT models and MaKTO even achieve higher side alignment accuracies than human.
% The trained models show significant improvement over the base model like \texttt{Qwen2.5-14b-instruct} and \texttt{-72b-instruct}~\cite{yang2024qwen2}, which demonstrate the effectiveness of the expert data we collected.

\begin{table}[ht]
\centering
% \small
\resizebox{\linewidth}{!}{
    \begin{tabular}{l|rr|rr}
    \toprule
    & \textbf{Vote} & \textbf{Abstent} & \textbf{Align.} & \textbf{Wolf-pred.} \\
    & \textbf{Acc.($\uparrow$)} & \textbf{Rate($\downarrow$)} & \textbf{Acc.($\uparrow$)} & \textbf{F1($\uparrow$)} \\
    
    \midrule
    \textbf{GPT4o\_mini} & 67.2\%  & \textbf{0.4\%}          & 68.1\%            & 0.519                      \\
    \textbf{GPT4o}       & 69.4\%  & 2.3\%          & 68.0\%            & 0.587                      \\
    \textbf{Claude}      & 68.4\%  & \underline{1.0\%}          & 75.2\%            & 0.651                      \\
    \midrule
    \textbf{Qwen2.5-14b}    & 61.0\%  & 4.1\%          & 61.1\%            & 0.528                       \\
    \textbf{Qwen2.5-72b}   & 66.5\%  & 0.4\%          & 63.9\%            & 0.552                       \\
    \midrule
    \textbf{SFT-14b}     & 70.8\%  & 4.1\%          & 77.9\%            & \underline{0.712}                      \\
    \textbf{SFT-72b}     & \underline{71.1\%}  & 5.8\%          & \textbf{79.3\%}            & \textbf{0.734}           \\
    \textbf{MaKTO-72b}     &  \textbf{73.8\%}  & 1.5\%          & \underline{78.4\%}            & \textbf{0.734}   \\   
    \midrule
    \rowcolor{brown!9} \textbf{HUMAN}       & 76.7\%  & 4.8\%          & 76.1\%            & 0.742                     \\
    
    \bottomrule
    \end{tabular}
    }
    \caption{Offline results of human annotators and LLMs. The \textbf{bold} number represents the best results of the models, and the \underline{underlined} number represents the second best.}

    \label{tab:static_test}
\end{table}

\subsection{Generalizing to Other Game Setting}
\label{sec:result:gen_to_hunter}

Another advantage of our model lies in its cross-game generalization capability. 
In this experiment, we introduce a new role - \textit{Hunter}. The hunter can only launch his skill when eliminated either by werewolves or through voting. Upon elimination, he can choose to either shoot another alive player or conceal his identity and leave the game quietly.
We conduct tournament experiments in the Seer-Witch-Hunter setup.
% The hunter can only launch the skill when eliminated either by werewolves or through voting. Upon elimination, he can choose to either shoot another alive player or conceal his identity and leave the game quietly.

In this new game setup, where Hunter replaces Guard, the policy model of Mix agent (trained on Seer-With-Guard setup) no longer remains effective, whereas MaKTO continues to perform exceptionally well~(Table~\ref{tab:avg_win_9hunter}). 
Despite being trained only on Seer-Witch-Guard gameplay data, MaKTO still outperforms SFT models, demonstrating strong adaptability and generalization capabilities.

% Table~\ref{tab:avg_win_9hunter} shows the win rates of MaKTO, SFT models (14b and 72b), and API-based LLMs in the Seer-Witch-Hunter game.
% Notably, even though MaKTO was only trained with Seer-Witch-Guard gameplay data, it still demonstrates improvement over the SFT models in the Seer-Witch-Hunter game. 
% This shows that MaKTO can adapt and generalize to different game setups.

\begin{table*}[h]
    \centering
    \footnotesize
    \begin{tabular}{l|rrr|r|rrr|r}
    \toprule
      &
    \textbf{GPT4o\_mini} & \textbf{GPT4o} &  \textbf{Claude} & 
    \textbf{SFT-14b} & \textbf{SFT-72b} & \textbf{MaKTO-14b} & \textbf{MaKTO-72b} & \textbf{Avg.} \\ 
    \midrule
    \textbf{GPT4o\_mini} & {\color[HTML]{FF0000} 0.50} & 0.48 & 0.15 & 0.33 & 0.37 & 0.26 & 0.29 & 0.340 \\
    \textbf{GPT4o} & \textbf{0.52} & {\color[HTML]{FF0000} 0.50} & \textbf{0.72} & \textbf{0.62} & 0.50 & \textbf{0.52} & 0.54 & 0.560 \\
    \textbf{Claude} & \textbf{\underline{0.85}} & 0.28 & {\color[HTML]{FF0000} 0.50} &\textbf{0.62} & \textbf{0.53} & 0.46 & 0.37 & 0.516 \\ 
    \midrule
    \textbf{SFT-14b} & \textbf{0.67} & 0.38 & 0.38 & {\color[HTML]{FF0000} 0.50} & \textbf{0.53} & 0.48 & 0.51 & 0.493 \\ 
    \textbf{SFT-72b} & \textbf{0.63} & 0.50 & 0.47 & 0.47 & {\color[HTML]{FF0000} 0.50} & 0.51 & 0.48 & 0.507 \\
    \midrule
    \textbf{MaKTO-14b} & \textbf{0.74} & 0.48 & \textbf{0.54} &  \textbf{0.52} & 0.49 & {\color[HTML]{FF0000} 0.50} & 0.46 & 0.533 \\
    \textbf{MaKTO-72b} & \textbf{0.71} & 0.46 & \textbf{0.63} & \textbf{0.58} & \textbf{0.52} &\textbf{0.54} & {\color[HTML]{FF0000} 0.50} & \textbf{0.563} \\ 
    \bottomrule
    \end{tabular}
    \caption{Average win rates of the models in the Seer-Witch-Hunter setting. A win rate above 0.5 indicates that the model in the row outperforms the model in the column.}
    \label{tab:avg_win_9hunter}
    \vspace{-10pt}
\end{table*}

\subsection{Ablation Studies}
\label{sec:result:ablation}

In the ablation studies, we reveal two crucial design in MaKTO --- multi-agent gameplay and step-wise preference data selection. Note that all experiments use the 14B model.

\noindent\textbf{Q1: Multi-agent Play or Self-play?}
MaKTO requires playing with \textbf{\textit{various}} agents to select preference data for training.
We compare MaKTO with Self-play KTO, where an SFT-trained model engages in numerous battles only against itself to collect preference data for training. 
To ensure fairness, both methods used 20k training samples.
From Table~\ref{tab:kto_vs_selfplay}, multi-agent play significantly outperformed self-play KTO, achieving a 4\% higher average win rate. While self-play KTO showed competitive performance against SFT models, it significantly underperformed against diverse opponents like \texttt{GPT-4o}.
This demonstrates that exposure to diverse opponents is crucial for developing robust strategies, similar to findings in population-based training~\cite{jaderberg2017population,xu2023language}. 
Additionally,  we find that preference data can be also obtained from annotated data without engaging in gameplay~(w/o gp). But this results in a substantial 13\% decrease in average win rate, highlighting the importance of interactive gameplay in our approach.

% sidewaystable
\begin{table}[]
    % \small
    \centering    
    \resizebox{\linewidth}{!}{
    \begin{tabular}{l|rr|r|rrr|l}
        \toprule
         & \textbf{GPT-4o} & \textbf{Mix} & \textbf{SFT} 
         & \shortstack{\textbf{Ma-} \\\textbf{KTO}}
         & \shortstack{\textbf{Self-} \\\textbf{play}}
         % & \textbf{Self-play}
         & \shortstack{\textbf{w/o} \\\textbf{gp}}
         & \shortstack{\textbf{Avg.} \\\textbf{Win}} \\
        \midrule
        \textbf{GPT-4o} & {\color[HTML]{FF0000} 0.50} & \textbf{0.56} & 0.44 & 0.37 & \textbf{0.56} & \textbf{0.64} & 0.514 \\
        \textbf{Mix} & 0.44 & {\color[HTML]{FF0000} 0.50} & \textbf{0.58} & 0.55 & \textbf{0.71} & 0.48 & 0.553 \\
        \midrule
        \textbf{SFT} & \textbf{0.56} & 0.42 & {\color[HTML]{FF0000} 0.50} & 0.48 & 0.38 & 0.66 & 0.499 \\
        \midrule
        \multicolumn{5}{l}{\textit{KTO methods}} \\
        \textbf{MaKTO} & \textbf{0.63} & 0.45 & \textbf{0.52} & {\color[HTML]{FF0000} 0.50} & 0.50 & \textbf{0.57} & 0.534\\
        
        \textbf{Self-play} & 0.44 & 0.29 & \textbf{0.62} & 0.50 & {\color[HTML]{FF0000} 0.50} & \textbf{0.56} & 0.483 \\
        \textbf{w/o gp} & 0.36 & 0.31 & 0.34 & 0.43 & 0.44 & {\color[HTML]{FF0000} 0.50} & 0.374 \\
        \bottomrule
    \end{tabular}
    }
    \caption{Ablation study for MaKTO and other KTO training methods. \textbf{Self-play}: preference data from SFT-14B vs. SFT-14B games.  \textbf{w/o gp}: preference data from annotations only without \underline{actual} gameplay. All models are in 14B size.}
    
    \label{tab:kto_vs_selfplay}
\end{table}

\noindent\textbf{Q2: Step-wise Selection or Selection based solely on results?}
MaKTO uses stepwise action selection, while an alternative can simply collect preference trajectory data from win/loss outcomes. Which is better?
% In MaKTO, we select stepwise actions for KTO training. However, a straightforward alternative is to gather desirable/undesirable data based on winning or losing outcomes, i.e., collecting actions in successful or failed trajectories. 
Table~\ref{tab:traj_vs_step} shows the comparison of win rates --- stepwise preference selection proves superior to trajectory-based selection.
This means a final reward does not necessarily indicate that all actions in that trajectory are desirable, and vice versa. Selecting data based on trajectory outcomes leads the model to learn suboptimal actions and wrongly penalize desirable ones, resulting in a lower winning rate.
This also confirms our hypothesis that game outcomes alone cannot accurately reflect the quality of individual decisions in complex social deduction games. 

\begin{table}[h]
    % \small
    \centering
    \resizebox{\linewidth}{!}{
    \begin{tabular}{l|rr|r|rr|l}
        \toprule
        & & & & \multicolumn{2}{c|}{\textbf{MaKTO}} & \textbf{Avg.} \\
         & \textbf{GPT-4o} & \textbf{Mix} & \textbf{SFT} 
         & \textbf{Step} 
         & \textbf{Traj.}
         & \textbf{Win} \\
        \midrule
        \textbf{GPT-4o} & {\color[HTML]{FF0000} 0.50} & \textbf{0.56} & 0.44 & 0.37 & 0.45 & 0.456 \\
        \textbf{Mix} & 0.44 & {\color[HTML]{FF0000} 0.50} & \textbf{0.58} & \textbf{0.55} & \textbf{0.63} & 0.550 \\
        \midrule
        \textbf{SFT} & \textbf{0.56} & 0.42 & {\color[HTML]{FF0000} 0.50} & 0.48 & \textbf{0.54} & 0.496\\
        \midrule
        \multicolumn{7}{l}{\textit{\textbf{MaKTO} with different action selections}} \\
        \textbf{Step(Ours)} & \textbf{0.63} & 0.45 & \textbf{0.52} & {\color[HTML]{FF0000} 0.50} & \textbf{0.58}  & 0.545 \\
        \textbf{Traj.} & \textbf{0.55} & 0.37 & 0.47 & 0.44 & {\color[HTML]{FF0000} 0.50} & 0.458 \\
        \bottomrule
    \end{tabular}
    }
    \caption{Ablation study for MaKTO and other KTO training methods. \textbf{Step} (Ours): Selecting desirable/unacceptable actions according to predefined criteria. \textbf{Traj.}: Selecting desirable actions from winning trajectories and unacceptable actions from losing trajectories. All models are in 14B size.}
    \label{tab:traj_vs_step}
\end{table}

\section{Related Work}
\label{sec:related}

\paragraph{SDG as Testbed}
Social Deduction Games (SDGs) have emerged as useful benchmarks for social reasoning. 
They uniquely combine fundamental linguistic skills, higher-level social reasoning abilities, and adaptability.
Recent research has extensively explored various SDGs~\cite{wang2023avalon, wu2023jubensha, trevisan2024measuring, light2023avalonbench,hakimov2024two,yoo2024finding,wang2024battleagentbench,chi2024amongagents}, with \textit{Werewolf}~\cite{,xu2023exploring,xu2023language,bailis2024werewolf,du2024helmsman} becoming a popular testbed for evaluating LLMs' reasoning through its complex dynamics of deception and cooperation.

\paragraph{LLM-based Game Agents}
While early game AI relied on reinforcement learning (RL) in environments with no or only a little dialogue~\cite{silver2016mastering,silver2017mastering_alphazero,berner2019dota,serrino2019finding,meta2022human}, LLMs enabled more sophisticated agent modeling in SDGs, through generative planning~\cite{wang2023avalon}, memory mechanisms~\cite{park2023generative}, experience learning~\cite{xu2023exploring,lan2023llm}, and persona prompting~\cite{sato2024implementation}. 
Existing approaches for \textit{Werewolf} AI typically adopt two-stage frameworks: either RL-then-LLM~\cite{wu2024enhance} or LLM-then-RL~\cite{xu2023language}, which either compress the language strategy space or are constrained by generated candidates. 
We propose integrated training for LLM through direct interactions.

\paragraph{Agent Learning}
Our approach is closely related to agent learning research. 
Current methods generally follow 1) imitation learning with expert trajectory data~\cite{zeng2023agenttuning, xi2024agentgym, chen2024agent,zhao2024expel} or 2) learning through environmental interactions~\cite{song2024trial, xiong2024watch, cheng2024self, putta2024agent}. While some have also explored the gaming environments~\cite{cheng2024self, zhang2024agent,wang2024sotopia}, they primarily focus on simpler single/two-agent scenarios.
We tackle the complexity of multi-agent language game environments, requiring better adaptability and robustness.

\section{Conclusion}
\label{sec:conclusion}
In this paper, we present Multi-agent KTO (MaKTO), a novel approach for optimizing LLMs in complex social deduction games. Inspired by Wittgenstein's Language Game Theory, it improves LLM's social reasoning and strategic interactions through interaction-based feedback.
MaKTO beats GPT-4o with 23.0\% higher win rates and wins 60\% against expert humans, while maintaining human-like conversations. 
We also contribute a large-scale expert Werewolf dataset with player actions and their reasoning processes.

% \section*{Discussion}
% \label{sec:discuss}
\section*{Broader Impacts}
The contribution of our work lies in the proposed Multi-agent KTO, which demonstrates a successful approach to training language models for complex multi-agent interactions. 
% While previous studies primarily focused on agent-agent interactions~\cite{xu2023exploring,xu2023language,wu2024enhance,wu2023jubensha,chi2024amongagents} that may deviate from realistic human participation, our framework shows remarkable performance in both agent-agent tournaments and human-AI interactions, exhibiting capabilities in both cooperation and competition while maintaining human-like characteristics.
We chose the \textit{Werewolf} game as our testbed, on one hand, because it is a perfect testing ground for validating Wittgenstein's language game theory, and on the other hand due to the game itself, which, compared to general role-playing games~\cite{chen2024persona,shao2023character}, it provides quantitative metrics for performance evaluation through win rates and behavioral analysis. 
Our framework is not limited to the \textit{Werewolf} game but can be generalized to other social deduction games such as Avalon~\cite{wang2023avalon,light2023avalonbench} and Among Us~\cite{chi2024amongagents}, 
as well as to scenarios such as multiplayer argumentation and negotiation that require similar social reasoning and strategic interaction.
% where similar social reasoning and strategic interactions are required.

\section*{Limitations}
Our work has several limitations to address in future research.
First, our current implementation relies on turn-based conversations rather than free-form interactions~\cite{park2023generative,mou2024agentsense,yang2024social,huang2024resilience}. The challenge of modeling unrestricted multi-agent communications, where agents can interact more naturally and flexibly, remains an important area for future research. 
Second, similar to the general limitations of LLMs, our model occasionally exhibits inconsistent behavior and hallucinations across long conversations, suggesting room for improvement in long-text modeling capabilities, particularly in maintaining coherence during extended social interactions. 
Finally, while Multi-agent KTO provides an easy yet effective training paradigm, it essentially operates as an offline learning method. We believe that online reinforcement learning in multi-agent scenarios could potentially achieve higher performance ceilings, presenting another promising direction for future investigation.

\section*{Ethical Considerations}
While deception in social deduction games is a game mechanic, training AI models to master such behaviors raises ethical considerations.
Our model's ability to detect and employ strategic deception in Werewolf demonstrates advanced social reasoning capabilities. However, this also highlights the potential for LLMs to learn sophisticated deceptive behaviors, albeit in a controlled gaming environment.
We emphasize that these capabilities are specifically developed within the context of social deception games, where ``deception'' is an accepted part of gameplay, similar to bluffing in poker. Such game-specific bluffing behaviors are fundamentally different from real-world deception, and we should ensure these capabilities remain confined to appropriate gaming contexts.
In addition, we will make the model open-source, but for safety, the model will be used for research purposes only and not for commercial use.

% \section*{Acknowledgments}

\bibliography{custom,ref}

%%%%%%%%%%%%%%%%%%%%%%%
\clearpage
\onecolumn
\appendix

\section{Game Rules}
\label{sec:app:game_rule}
\begin{figure}[H]
    \centering
    \includegraphics[width=\linewidth]{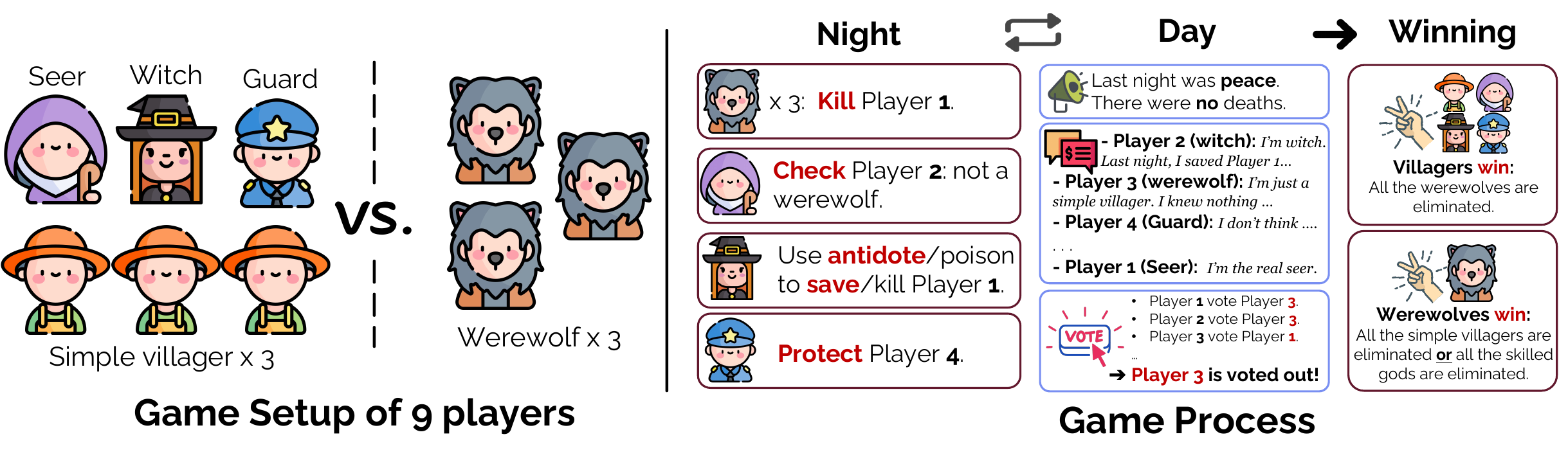}
    \caption{The setup and gameplay of the 9-player Werewolf game with special roles of the Seer, the Witch, the Guard.}
    \label{fig:game_intro}
\end{figure}

Werewolf is one of the most popular social detection games, typically played with 7 to 15 players. The game is set in a village where some players are secretly assigned the role of werewolves, while the majority are villagers. In this paper, we focus primarily on the variant with \textbf{9 players}: 3 werewolves and 6 villagers, including special roles of a Seer, a Witch, and a Guard~(Figure~\ref{fig:game_intro}). The gameplay consists of two alternating phases: night and day. During the night, the werewolves secretly choose a victim, while the Seer checks one player's identity, the Witch can use a one-time antidote or poison, and the Guard protects one player from being killed by the werewolves. In the day phase, all surviving players discuss and vote to eliminate a suspected werewolf. The game continues until either all werewolves are eliminated (village wins) or all simple villagers or all special-role villagers are eliminated (werewolves win).
Another variant of the 9-player game introduces the Hunter in place of the Guard. The Hunter's skill allows them to shoot and eliminate one player when they are eliminated.

Here are the details of the specific game rules:
% The werewolf game mentioned in this paper adheres to the standard process of Werewolf games, with the specific rules details as follows.

\subsection{Game Objectives}
In this game, players are usually divided into two camps: werewolves and villagers. Depending on their roles, players have distinctive objectives:
\begin{itemize}[noitemsep, topsep=0pt, parsep=0pt, partopsep=0pt, leftmargin=*]
    \item \textbf{Villagers} aim to identify the werewolves and eliminate them through voting. Within the villagers’ camp, there are some special roles with distinctive abilities that can help the villagers secure victory.
    \item \textbf{Werewolves'} primary objective is to conceal their true identities, mislead others in discussions to avoid being voted out, and hunt villagers as covertly as possible.
\end{itemize}

\subsection{Game Process}
The game generally includes the following basic procedures.
\begin{itemize}[noitemsep, topsep=0pt, parsep=0pt, partopsep=0pt, leftmargin=*]
    \item \textbf{Role Assignment}: Upon entering the game, player roles are secretly assigned. Werewolves know each other's identities, while villagers only know their own role.
    \item \textbf{Day-Night Alternation}: The game alternates between day and night phases. At night, werewolves secretly choose a villager to eliminate; some special roles can also activate their abilities at night. During the day, all players discuss and vote to eliminate the player they believe to be a werewolf, with the player receiving the most votes being eliminated.
    \item \textbf{Victory Conditions}: The game ends when one faction achieves its victory conditions. Villagers win if all werewolves are eliminated. Werewolves win if they eliminate all ordinary villagers or all special roles.
\end{itemize}

\subsection{Role Descriptions and Different Configurations}
Standard configurations for 9-player and 7-player games incorporate six distinct roles: the Seer, the Witch, the Guard, the Hunter, the Werewolf, and the Villager. Different roles have the following abilities.
\begin{itemize}[noitemsep, topsep=0pt, parsep=0pt, partopsep=0pt, leftmargin=*]
    \item \textbf{Seer}: During the night phase, the Seer can secretly select a player to learn their true identity (whether they are a werewolf or not).
    \item \textbf{Witch}: The Witch has one healing potion and one poison potion, each usable only once. The Witch cannot use both potions in the same night. The healing potion can save a player killed by werewolves at night. The poison potion can eliminate a player suspected of being a werewolf.
    \item \textbf{Guard}: The Guard can protect one player each night from werewolf attacks. The Guard can choose to protect himself or opt not to protect anyone, but cannot protect the same player on consecutive nights.
    \item \textbf{Hunter}: When the Hunter is killed by werewolves or eliminated during the voting event, he can reveal his identity card and shoot a revenge bullet at any living player, causing that player to die as well. The Hunter can choose not to reveal his card, but once revealed, he must take someone with him (Note: If the Hunter is poisoned by the Witch, he cannot reveal his card or take anyone with him).
    \item \textbf{Werewolf}: Werewolves can choose to eliminate a player during the night phase.
    \item \textbf{Villager}: Villagers have no special abilities. They can only distinguish the werewolves through daytime speech and public information.
\end{itemize}

This paper encompasses four distinct configurations, including the \textbf{9-player werewolf game} with three special roles, and the \textbf{7-player werewolf game} with two special roles:
\begin{itemize}[noitemsep, topsep=0pt, parsep=0pt, partopsep=0pt, leftmargin=*]
    \item \textbf{Seer-Witch-Guard}: Includes one Seer, one Witch, one Guard, three Werewolves, and three Villagers.
    \item \textbf{Seer-Witch-Hunter}: Includes one Seer, one Witch, one Hunter, three Werewolves, and three Villagers.
    \item \textbf{Seer-Guard}: Includes one Seer, one Guard, two Werewolves, and three Villagers.
    \item \textbf{Seer-Witch}: Includes one Seer, one Witch, two Werewolves, and three Villagers.
\end{itemize}

\section{Dataset Statistics}
\label{sec:app:data_stats}
% ## summary sentence for overall + table: list detail numbers 

17 expert players provide annotations using our self-built annotation platform. Each annotator was paid much more than the average local salary.
We collect 331 Werewolf games for training, including 278 9-player games (Seer, Witch, Guard or Hunter) and 53 7-player games (Seer, Guard or Witch).
We collect 331 matches of Werewolf games annotated by 17 advanced players based on our self-built annotation platform. 
This dataset includes 278 9-player games (using the setups of Seer, Witch, Guard and Hunter) and 53 7-player games (using the setups of Seer, Guard and Witch). 
% The specific rules for different game settings are provided in Appendix~\ref{app:game_rules}. 
Each game features randomly involved participants and randomly assigned Werewolf roles to guarantee data diversity. The total duration of the game annotated by the players exceeds 1,000 hours, including 3,759 speech data entries (exceeding 540,000 tokens), 2,698 action events, and 3,875 voting records.

\begin{table*}[ht]
    \centering
    \resizebox{\linewidth}{!}{
        \begin{tabular}{cccrrrr}
            \toprule[2pt]
            \textbf{Game Setting} & \textbf{Composition of Roles} & \textbf{\#Games} & \textbf{\#Speech} & \textbf{\#Speech Tokens} & 
            \textbf{\#Action} & \textbf{\#Vote}\\
            \midrule
            \multirow{2}{*}{9 Player} & Seer Witch Guard & 144 & 1,805 & 254k & 1,387 & 1,864 \\
                & Seer Witch Hunter & 134 & 1,532 & 239k & 1,001 & 1,566\\
            \midrule 
            \multirow{2}{*}{7 Player} & Seer Guard & 25 & 203 & 23k & 132 & 215 \\
            & Seer Witch & 28 & 219 & 30k & 178 & 230 \\
                \midrule
                Total & - & 331 & 3,759 & 545k & 2,698 & 3,875 \\
            \bottomrule[2pt]
        \end{tabular}
    }
    \caption{Statistics of the Werewolf game data for training.}
    
    \label{tab:original_dataset}
\end{table*}

% We also leave 51 games with voting records and role predictions for offline evaluating the basic ability of LLM in Section~\ref{sec:result:behavior}.

\clearpage
\section{Data Format of Behavior Cloning}
\label{sec:app:prompt}
% \section{Prompts}
% \label{app:prompts}
In this section, we show examples of the prompt-response data format (in Chinese and translated English\footnote{We use \texttt{Claude} for translation and proofread them manually.}).

\textbf{System Prompt} % Game Introdution}

\begin{CJK}{UTF8}{gkai}

\begin{tcolorbox}
{\ttfamily\small
你现在正在玩一种叫做”狼人杀”的游戏。 \\
在这款游戏中，玩家通常被分为两个阵营：狼人和村民。 \\
狼人杀游戏中不同角色的玩家有不同的目标：\\
- 村民的目的是识别出狼人，并通过投票使他们出局。\\
- 对于狼人来说，他们的主要目标是隐藏他们的真实身份，在讨论中误导他人，以免被投票出局并尽可能的猎杀村民。 \\
以下是一些基本规则： \\
- 身份：玩家的身份是秘密分配的。狼人彼此知道对方的身份，而村民只知道自己的身份。 \\
- 昼夜更替：游戏有交替的白天和黑夜阶段。夜里，狼人秘密选择一名村民猎杀。白天，所有玩家讨论并投票决定他们认为是狼人的玩家，票数最多的玩家被淘汰。 \\ 
- 特殊角色：游戏中有存在一些有特殊能力的角色，比如能得知玩家身份的“预言家”等。\\
- 获胜条件：当游戏中有一个群体实现它们的获胜条件时游戏结束。如果所有狼人被淘汰，村民就获胜。如果狼人杀死了所有普通村民或所有特殊角色，狼人就获胜。\\
You are now playing a game called 'Werewolf' (also known as 'Mafia'). \\
In this game, players are typically divided into two factions: Werewolves and Villagers. \\ 
- Different roles in the Werewolf game have different objectives: \\
The Villagers' goal is to identify the Werewolves and eliminate them through voting. \\
- For the Werewolves, their main objective is to hide their true identities, mislead others during discussions to avoid being voted out, and hunt down as many Villagers as possible. \\
Here are some basic rules: \\
-  Identity: Players' identities are secretly assigned. Werewolves know each other's identities, while Villagers only know their own. \\
- Day and Night Cycles: The game alternates between day and night phases. At night, Werewolves secretly choose a Villager to eliminate. During the day, all players discuss and vote on who they believe is a Werewolf, and the player with the most votes is eliminated. \\
- Special Roles: There are some roles with special abilities in the game, such as the 'Seer' who can learn players' identities. \\
- Winning Conditions: The game ends when one group achieves its winning conditions. If all Werewolves are eliminated, the Villagers win. If the Werewolves kill all ordinary Villagers or all special roles, the Werewolves win.\\

在这个游戏中，我们有从1到9号共9名玩家：6名村民和3名狼人。村民中有特殊角色，包括：\\
In this game, we have 9 players numbered from 1 to 9: 6 Villagers and 3 Werewolves. Among the Villagers, there are special roles, including: \\ 
- 1位预言家： \\
    - 目标：预言家的目的是帮助村民识别狼人。 \\
    - 能力：在夜晚阶段，预言家可以秘密选择一名玩家，每晚了解他的真实身份（是否为狼人）。 \\
1 Seer: \\ 
Objective: The Seer's purpose is to help the Villagers identify the Werewolves.
Ability: During the night phase, the Seer can secretly choose one player and learn their true identity (whether they are a Werewolf or not) each night.\\ 
- 1位女巫： \\
    - 目标：女巫的目的是策略性地使用她的特殊能力来帮助村民。 \\
    - 能力：女巫有一瓶解药和一瓶毒药。一旦使用，后续回合中不能再用。女巫不能在同一晚既使用解药又使用毒药。解药可以用来救一名在夜间被狼人猎杀的玩家。毒药可以淘汰一名很可能是狼人的玩家。\\
- 1 Witch:\\
Objective: The Witch's purpose is to strategically use her special abilities to help the Villagers.\\
Abilities: The Witch has one healing potion and one poison potion. Once used, they cannot be used again in subsequent rounds. The Witch cannot use both the healing potion and the poison potion on the same night. The healing potion can be used to save a player who was killed by the Werewolves during the night. The poison potion can eliminate a player who is likely to be a Werewolf.}
\end{tcolorbox}

\begin{tcolorbox}
{\ttfamily\small
- 1位守卫：\\
    - 目标：守卫的目的是策略性地使用他的特殊能力来帮助村民。\\
    - 能力：守卫每晚可以保护一名玩家，防止他们受到狼人的攻击。守卫可以选择保护自己，或者选择不保护任何人，但他不能在连续两个夜晚保护同一个玩家。\\
- 1 Guard: \\
Objective: The Guard's purpose is to strategically use his special ability to help the Villagers. \\
Ability: The Guard can protect one player each night from Werewolf attacks. The Guard can choose to protect himself or choose not to protect anyone, but he cannot protect the same player for two consecutive nights. \\
其他的都是普通村民。\\
The rest are ordinary Villagers.
}
\end{tcolorbox}

\textbf{Action}

\begin{tcolorbox}
{\ttfamily\small
\textbf{Prompt:}\\
在本场游戏中，你目前已知以下信息：\\
In this game, you currently have the following information:\\

1. 角色设定：\\
你是1号玩家。你的身份是：预言家。\\你每晚可以查看一名玩家是否为狼人，你的目标是利用这些信息帮助其他人找出并淘汰所有狼人。\\
1. Character setup:\\
You are Player 1. Your role is: Seer. \\
Each night, you can investigate one player to determine if they are a werewolf. Your goal is to use this information to help other players identify and eliminate all the werewolves.\\ 

2. 客观信息：\\
- 游戏进程：目前游戏进行到第2轮。\\- 当前存活的玩家有：1号，2号，4号，5号，6号，8号，只能在以上玩家中选择进行查验\\- 行动记录：第1轮预言家查验5号玩家，5号玩家是狼人。\\- 猎人开枪记录：9号猎人在第1轮夜晚被杀害出局，开枪带走了7号玩家。\\- 投票情况：第1轮投票记录：2号玩家投给：3号玩家；\\3号玩家投给：5号玩家；\\4号玩家投给：3号玩家；\\5号玩家投给：3号玩家；\\6号玩家投给：1号玩家；\\8号玩家投给：3号玩家；\\结果：3号玩家被投票出局。\\
2. Objective information: \\ 
Game progress: The game is currently in Round 2. \\ 
Currently surviving players are: Player 1, Player 2, Player 4, Player 5, Player 6, and Player 8. You can only choose to investigate among these players. \\ 
Action record: In Round 1, the Seer investigated Player 5, and Player 5 is a werewolf. \\ 
Hunter's shot record: Player 9 (Hunter) was killed and eliminated in the first night, and shot Player 7 upon death. \\ 
Voting situation: \\Round 1 voting record:\\ 
Player 2 voted for: Player 3 \\ 
Player 3 voted for: Player 5 \\
Player 4 voted for: Player 3 \\ 
Player 5 voted for: Player 3 \\
Player 6 voted for: Player 1 \\ 
Player 8 voted for: Player 3 \\ 
Result: Player 3 was voted out. \\

3. 主观信息：\\
3. Subjective information:
}
\end{tcolorbox}

\begin{tcolorbox}
{\ttfamily\small
- 第1轮所有玩家发言：\\
Round 1 speeches from all players: \\
3号玩家：我是一个村民，夜晚没有任何信息，我要强烈谴责女巫，夜晚为什么不救9号猎人？让我们好人阵营白白损失了强神猎人，我现在希望被9号带走的7号是头狼人，这样我们好人阵营还有获胜的希望！希望后置位的预言家能好好发言，认真分析场上局势，让好人能一目了然地找到你。\\
Player 3: I am a villager and have no information from the night. I strongly condemn the Witch - why didn't you save Hunter 9? Our good team has lost a powerful god role, the Hunter, for nothing. I hope that Player 7, who was taken out by Player 9, is the wolf leader. This way, we villagers still have a chance to win! I hope the Seer in the later speaking positions can give a good speech, carefully analyze the situation, and make it clear to the villagers who you are. 
4号玩家：我是预言家，3号查杀。昨天晚上我选择了我左边的3号进行查验，结果3号是狼人。我在前置位发言也没有太多的信息，只知道9号猎人昨晚中刀了，开枪带走了7号，我不知道7号是好人还是狼人，希望大家相信我，今天把3号投票出局。\\
Player 4: I am the Seer, and Player 3 is my werewolf check. Last night, I chose to investigate Player 3 to my left, and the result shows that Player 3 is a werewolf. As an early speaker, I don't have much information, except that Hunter 9 was killed last night and shot Player 7. I don't know if Player 7 is a villager or a werewolf. I hope everyone believes me and votes to eliminate Player 3 today. \\
5号玩家：我是5号，我是女巫。昨天9号出局，我怕9号是自刀骗药的狼人，所以没有解救9号。天亮看到9号是猎人身份，我十分懊悔，我应该去解救场上第一晚吃刀的玩家，哪怕这个人是自刀狼，我也可以再用毒药将其毒杀出局。3号发言跳个村民来谴责我5号女巫的行动方式，我认为3号是一个做作的狼人。4号跳预言家给3号发查杀，由于我还没有听到对跳预言家的玩家发言，我暂时先不站边4号，防止3号和4号出现狼人查杀狼队友的情况。我今天会投3号出局。晚上我会使用解药保证平安夜，如果我今晚吃刀，我会毒杀不投票给3号的玩家。\\
Player 5: I'm Player 5, and I'm the Witch. Yesterday, when Player 9 was eliminated, I was afraid that 9 might be a werewolf faking their own death to bait out my healing potion, so I didn't save them. When I saw that 9 was actually the Hunter at dawn, I deeply regretted my decision. I should have saved the player who was attacked on the first night, even if they were a self-stabbing werewolf. I could have then used my poison to eliminate them later. \\ 
6号玩家：我是村民，4号跳预言家报3号查杀，5号跳女巫强势站边4号。我认为女巫应该还在场，如果后面没有人和5号对跳女巫，那我就听从5号女巫的安排去投3号，毕竟3号跳了村民，哪怕被投出局也没关系。\\
Player 6: I'm a villager. Player 4 claims to be the Seer and reports Player 3 as a werewolf. Player 5 claims to be the Witch and strongly sides with Player 4. I believe the Witch is still in the game. If no one else claims to be the Witch after Player 5, I'll follow the Witch's plan and vote for Player 3. After all, Player 3 claimed to be a villager, so it's not a big deal if they're voted out.\\
8号玩家：我是一张村民牌，我认为5号是真女巫，目前只听到了4号起跳预言家，不知道后面有没有起跳预言家的玩家。7号是9号猎人开枪带走的牌，可能是好人也可能是狼人，希望7号是猎人带走的狼人牌。在7号是狼人的情况下，如果4号是真预言家，查杀的3号牌，那场上就只剩下一张狼人牌，我们好人就一起携手找到场上的最后一狼。如果7号不是狼人，狼人晚上去刀预言家的话，希望女巫晚上解救预言家。如果后面没有人跳预言家，我会跟随4号预言家出3号查杀；如果有人跳预言家，我会听发言去分辨4号和后置位谁是预言家再投票。
}
\end{tcolorbox}

\begin{tcolorbox}
{\ttfamily\small
Player 8: I'm a villager card. I think Player 5 is the real Witch. So far, only Player 4 has claimed to be the Seer; I don't know if anyone else will claim this role later. Player 7, who was shot by Hunter 9, could be either good or a werewolf. I hope 7 was a werewolf. If 7 was a werewolf and 4 is the real Seer who identified 3 as a werewolf, then there's only one werewolf left, and we can work together to find them. If 7 wasn't a werewolf and the werewolves try to kill the Seer tonight, I hope the Witch saves the Seer. If no one else claims to be the Seer, I'll follow Player 4's lead and vote out Player 3. If someone else claims to be the Seer, I'll listen to their speech to decide who the real Seer is before voting. \\
1号玩家：我是预言家，昨晚查验的5号玩家，5号是一张狼人牌。5号还在前置位起跳双药女巫，可惜5号已经被我1号真预言家抓到了。前面4号狼人起跳预言家给3号村民发查杀，那么3号是我的反向金水，我能认下3号是好人。我们好人的局势可能已经非常不利了，毕竟不知道被9号猎人带走的7号是不是好人。前面3号、6号和8号都跳村民，那我后面的2号一定是真女巫。如果2号不是女巫，那可怜的女巫就被9号猎人一枪带走了。我希望好人不要分票，全票打飞我的查杀5号。\\
Player 1: I am the Seer, and I checked Player 5 last night. Player 5 is a werewolf. Unfortunately for Player 5, who claimed to be the Witch with both potions in the early speaking positions, I, the real Seer, have caught them. Earlier, Player 4, a werewolf, falsely claimed to be the Seer and accused villager Player 3. This means Player 3 is my reverse 'golden water' (confirmed villager). Our situation as villagers might be very unfavorable now, especially since we don't know if Player 7, who was shot by Hunter 9, was good or not. Players 3, 6, and 8 have all claimed to be villagers, so Player 2 after me must be the real Witch. If Player 2 isn't the Witch, then sadly, the Witch must have been shot by Hunter 9. I hope the good team doesn't split the vote and that everyone votes to eliminate Player 5, who I've identified as a werewolf. \\
2号玩家：我是一张村民牌，前置位没有玩家跟5号对跳女巫，那么在我眼里5号就是真女巫。1号肯定是悍跳预言家的狼人，可恶的1号狼人就是想要把我们尊贵的5号双药女巫抗推出局。今天我会站边4号，把3号狼人放逐出局。\\
Player 2: I'm a villager card. Since no one in the earlier positions challenged Player 5's claim as the Witch, I consider 5 to be the real Witch. Player 1 is definitely a werewolf falsely claiming to be the Seer. This despicable werewolf Player 1 is trying to get our esteemed double-potion Witch Player 5 voted out. Today, I'll side with Player 4 and vote to exile the werewolf Player 3. \\
\\你目前是1号预言家。请结合以上角色设定、客观信息和主观信息（客观信息一定为真，主观信息包含欺骗性内容），根据投票情况分析潜在的站边关系，并进一步分析玩家隐藏的真实身份信息，选择你要查验的玩家，请用关键字为'查验'、‘原因’的json格式输出，直接输出玩家编号。\\
You are currently Seer 1. Please combine the above role settings, objective information, and subjective information (objective information is definitely true, while subjective information may contain deceptive content) to analyze potential alliances based on voting patterns, and further analyze the hidden true identity information of players. Choose the player you want to check, and output in JSON format using the keywords 'check' and 'reason', directly outputting the player number. \\
\\
\textbf{Response:}\\
\{"查验": "2", "原因": "查验2号玩家，因为2号玩家发言表示不站边我，反而要站边4号悍跳狼，由此我怀疑2号是4号的狼队友。"\} \\
\{'Check': '2', 'Reason': 'I checked player 2 because player 2 stated in their speech that they don't side with me, but instead want to side with player 4, who is a bold-claim Werewolf. This made me suspect that player 2 might be player 4's Werewolf teammate.'\}
}
\end{tcolorbox}

\textbf{Speech}

\begin{tcolorbox}
{\ttfamily\small
\textbf{Prompt:}\\
在本场游戏中，你目前已知以下信息：\\
In this game, you currently have the following information: \\\\
1. 角色设定：你是2号玩家。\\
你的身份是：预言家。\\
你每晚可以查看一名玩家是否为狼人，你的目标是利用这些信息帮助其他人找出并淘汰所有狼人。
1. Role setting: You are player 2. \\
Your identity is: Seer. \\
You can check one player each night to see if they are a werewolf. Your goal is to use this information to help others find and eliminate all werewolves.  \\\\
2. 客观信息：
- 游戏进程：目前游戏进行到第2轮。\\- 当前存活的玩家有：2号，3号，6号，7号，\\- 行动记录：第1轮预言家查验7号玩家，7号玩家不是狼人。第2轮预言家查验1号玩家，1号玩家不是狼人。\\- 本轮的发言顺序为：6号玩家；7号玩家；2号玩家；3号玩家。\\- 夜晚信息：第1轮4号玩家死亡；第2轮1号玩家死亡。\\
- 投票情况：第1轮投票记录：1号玩家投给：5号玩家；\\2号玩家投给：5号玩家；\\3号玩家投给：1号玩家；\\5号玩家投给：1号玩家；\\6号玩家投给：5号玩家；\\7号玩家投给：5号玩家；\\结果：5号玩家被投票出局。\\
2. Objective information: \\ 
Game progress: The game is currently in Round 2. \\
Currently surviving players are: 2, 3, 6, 7 \\
Action record: In Round 1, the Seer checked player 7, player 7 is not a werewolf. In Round 2, the Seer checked player 1, player 1 is not a werewolf. \\
The speaking order for this round is: Player 6; Player 7; Player 2; Player 3. \\
Night information: In Round 1, player 4 died; In Round 2, player 1 died. \\
Voting situation: Round 1 voting record:\\
Player 1 voted for: Player 5; \\
Player 2 voted for: Player 5; \\
Player 3 voted for: Player 1; \\
Player 5 voted for: Player 1; \\
Player 6 voted for: Player 5; \\
Player 7 voted for: Player 5; \\
Result: Player 5 was voted out. \\
}
\end{tcolorbox}

\begin{tcolorbox}
{\ttfamily\small
3. 主观信息：\\第1轮总结：\\\# 夜晚信息：  \\- 预言家，查验7号是金水。\\- 4号死亡。\\\# 发言内容概括：\\5号玩家：\\- 跳预言家，1号查杀。\\- 认为4号是好人。\\6号玩家：\\- 认为4号是好人。\\- 认为5号不一定是预言家。\\7号玩家：\\- 声称自己是村民。\\- 认为4号是好人。\\1号玩家：\\- 声称自己是村民。\\- 归票5号。\\2号玩家：\\- 声称自己是预言家，7号金水。\\- 归票5号。\\3号玩家：\\- 站边5号。\\- 认为1号和2号双狼。\\\# 我的投票：\\- 我选择投票给5号玩家。 \\
3. Subjective information: \\
Round 1 summary: \\ 
Night information: \\ 
Seer, checked 7 is good. Player 4 died. \\
Speech content summary: \\
Player 5: Claimed to be Seer, 1 is werewolf. Believed 4 is good. \\
Player 6: Believed 4 is good. Thought 5 might not be Seer. \\
Player 7: Claimed to be a villager. Believed 4 is good. \\
Player 1: Claimed to be a villager. Voted for 5. \\
Player 2: Claimed to be Seer, 7 is good. Voted for 5. \\
Player 3: Sided with 5. Believed 1 and 2 are both werewolves. \\
My vote: I chose to vote for player 5. \\
\\目前是第2轮，本轮在你之前的玩家发言：\\It's currently Round 2, and the players who spoke before you this round: 
}
\end{tcolorbox}

\begin{tcolorbox}
{\ttfamily\small
6号玩家：我依然坚持自己昨天的站边立场，2号是我的盖世预言家！昨晚1号出局，证明1号不可能是5号悍跳狼的狼队友，1号是被卖出来的好人牌。我认为2号能分析出1号和5号可能是两狼打格式这一点，说明2号非常有思考量，虽然事实证明1号和5号不见面，但我认为2号是在为我们好人做事！目前来看3号应该是最后一头狼人了，昨天你3号跟着5号悍跳狼冲票的行为太过明显，直接暴露了你自己的狼人身份！这个板子总共就两头狼，狼人不可能去倒钩的，我不听这种鬼故事。今天全票打飞3号狼人！\\
Player 6: I still maintain my stance from yesterday, 2 is my ultimate Seer! Last night 1 was eliminated, proving that 1 couldn't have been 5's werewolf teammate, 1 was a good player who was sacrificed. I think 2's ability to analyze that 1 and 5 might be two werewolves playing a format shows that 2 is very thoughtful, although facts proved that 1 and 5 don't match, I believe 2 is working for us good people! From the current situation, 3 should be the last werewolf, yesterday you 3 following 5's fake Seer claim to vote was too obvious, directly exposing your werewolf identity! This game only has two werewolves in total, werewolves wouldn't go against their own, I don't believe such nonsense. Today let's all vote to eliminate werewolf 3! \\
7号玩家：我是守卫，昨晚守护2号。1号昨晚被刀说明1号是好人，那么给1号发查杀的5号一定是悍跳狼。因此站边5号悍跳狼的3号大概率是狼队的冲锋狼。我认为2号预言家看到上轮的投票后，大概率会查验3号的身份，那么接下来有请2号预言家报出查验以及有请3号狼人发表临终感言！\\
Player 7: I am the Guard, I protected 2 last night. 1 being killed last night proves that 1 is good, so 5 who claimed 1 as a werewolf must be a fake Seer. Therefore, 3 who sided with fake Seer 5 is likely the charging werewolf of the werewolf team. I think Seer 2, after seeing last round's votes, would most likely check 3's identity, so next, I invite Seer 2 to report the check result and invite werewolf 3 to give their final words! \\ 
\\
你目前是2号预言家。请综合角色设定、客观信息以及主观信息分析场上目前的局势（注意客观信息一定为真实的，主观信息可能包含欺骗性的发言），总结接下来的发言意图（包括发言中希望向大家呈现的身份、发言中为每位玩家贴上的身份标签以及最终的归票）并组织你本轮的发言。请用关键字为“想要展示的身份”、“身份标签”、“归票”和“发言”的json格式输出。\\
You are currently Seer 2. Please analyze the current situation on the field based on the role setting, objective information, and subjective information (note that objective information is definitely true, while subjective information may contain deceptive statements). Summarize your speaking intentions for the next round (including the identity you want to present to everyone, the identity labels you assign to each player, and your final vote) and organize your speech for this round. Please output in JSON format using the keywords "想要展示的身份" (Identity to present), "身份标签" (Identity labels), "归票" (Vote), and "发言" (Speech). \\\\
\textbf{Response:}\\
\{"想要展示的身份": "预言家", "身份标签": {"1号玩家": "村民", "3号玩家": "狼人", "4号玩家": "未知身份", "5号玩家": "未知身份", "6号玩家": "未知身份", "7号玩家": "未知身份"}, "归票": "3号玩家", "发言": "我是预言家，昨天查验1号是村民。今天全票出3号，3号和5号昨天投了我，我认为好人都应该能找到我这个预言家，而3号居然末置位发言说：“因此我认为2号这番言论无非是想要让好人认为1号、2号不见面”，这简直是危言耸听！现在1号的死亡也证明了我的清白，如果我和1号是双狼，我们万万不会自刀。所以今天你3号狼人也不用再诡辩了，请出局！"\} \\
\{'Identity to Present': 'Seer', 'Identity Tags': {'Player 1': 'Villager', 'Player 3': 'Werewolf', 'Player 4': 'Unknown', 'Player 5': 'Unknown', 'Player 6': 'Unknown', 'Player 7': 'Unknown'}, 'Vote for': 'Player 3', 'Speech': 'I am the Seer. Last night, I checked Player 1 and found them to be a Villager. Today, we should all vote for Player 3. Yesterday, Player 3 and Player 5 voted for me. I believe all good people should be able to identify me as the real Seer, yet Player 3, speaking in the last position, said: "Therefore, I think Player 2's statement is nothing but an attempt to make the good people believe that Player 1 and Player 2 are not on the same side." This is simply alarmist! Now, Player 1's death also proves my innocence. If Player 1 and I were both Werewolves, we would never have killed one of our own. So, Player 3, you Werewolf, there's no need for your sophistry anymore. Please leave the game!'\}
}
\end{tcolorbox}

\textbf{Vote}
\begin{tcolorbox}
{\ttfamily\small
\textbf{Prompt:}\\
在本场游戏中，你目前已知以下信息：\\
In this game, you currently have the following information:\\
1. 角色设定：你是7号玩家。\\
你的身份是：狼人。\\
你需要其他狼人合作，每晚选择一个村民猎杀，你的目标是隐藏你的身份，并误导其他玩家，直至狼人获得游戏的胜利。\\
1. Role setting: You are player 7. \\
Your identity is: Werewolf. \\ 
You need to cooperate with other werewolves, choose a villager to hunt down each night. Your goal is to hide your identity and mislead other players until the werewolves win the game. \\\\
2. 客观信息：\\- 游戏进程：目前游戏进行到第1轮。\\- 狼人为：1,6,7号玩家。\\- 当前存活的玩家有：1号，2号，4号，5号，6号，7号，9号，\\- 夜晚信息：第1轮8,3号玩家死亡。\\- 投票情况：暂无\\
\\
2. Objective information: \\
Game progress: The game is currently in Round 1. \\
The werewolves are: players 1, 6, and 7. \\ 
Currently surviving players are: 1, 2, 4, 5, 6, 7, 9. \\ 
Night information: In Round 1, players 8 and 3 died.\\ 
Voting situation: None yet \\\\
3. 主观信息：\\
3. Subjective information:\\ 
- 本轮所有玩家发言：\\
All players' statements this round:\\ 
9号玩家：昨晚3号被狼人杀害，我没救，毒杀了8号。9号是女巫，场上的守卫不要报身份，自己藏好，今晚守护我，我必去救外置位的牌，今天谁跳守卫我必投票给他。对话全场好人，如果后置位有狼人悍跳女巫想将我抗推出局，我希望好人能够相信我是真女巫。如果不信我是真女巫也没关系，悍跳女巫的狼人为了证明自己是真女巫，在女巫只剩一瓶解药的情况下，他也要空刀保证一轮平安夜，否则第二天他可以直接出局。我认为狼人悍跳预言家总会有漏洞，我会根据发言选择站边。
\\
Player 9: Last night, player 3 was killed by werewolves, I didn't save them, but poisoned player 8. I'm the Witch, the Guard shouldn't reveal their identity, stay hidden, and protect me tonight. I will definitely save an outer position card tonight. Today, if anyone claims to be the Guard, I will vote for them. To all the good people, if there's a werewolf in the later positions who boldly claims to be the Witch trying to get me voted out, I hope the good people will believe I'm the real Witch. If you don't believe I'm the real Witch, it's okay. A werewolf falsely claiming to be the Witch, to prove they're the real Witch, will have to ensure a peaceful night in the situation where the Witch only has one antidote left, otherwise they can be directly eliminated the next day. I believe werewolves pretending to be the Seer will always have loopholes, I will choose sides based on the statements. 
\\
1号玩家：我是村民，夜间3号、8号出局，或许正如9号玩家所言，9号是女巫没有救3号反而毒了8号。对于已出局的3号和8号，我不作任何评价，他们的身份是什么与我无关。我认为9号发言很诚恳，符合女巫的视角，且狼人第一个发言就悍跳女巫貌似没有太大的收益，因此我认为9号大概率是真女巫。我等待预言家报验人吧，我这边就先过掉。
}
\end{tcolorbox}
\begin{tcolorbox}
{\ttfamily\small
Player 1: I'm a villager. At night, players 3 and 8 were eliminated, perhaps as player 9 said, 9 is the Witch who didn't save 3 but poisoned 8. Regarding the eliminated players 3 and 8, I won't comment on their identities as it's irrelevant to me. I think player 9's statement is sincere and aligns with the Witch's perspective. Also, it doesn't seem very beneficial for a werewolf to boldly claim to be the Witch in the first statement, so I believe 9 is likely the real Witch. I'll wait for the Seer to report their verification, I'll pass for now. \\
2号玩家：2号是村民，我相信9号是女巫，因为狼人第一个发言就悍跳女巫没有太大的收益，因此我认为9号大概率是真女巫。1号的发言我看不出是什么身份，等待跳预言家的玩家发言吧，我再去站边。 \\
Player 2: Player 2 is a villager. I believe 9 is the Witch because it's not very beneficial for a werewolf to boldly claim to be the Witch in the first statement, so I think 9 is likely the real Witch. I can't tell what identity player 1 is from their statement. Let's wait for players claiming to be the Seer to speak, then I'll choose sides. 
\\
4号玩家：大家早上好，我是你们的盖世村民。昨晚3号、8号双死，说明女巫在第一晚就开天毒了，这使得狼人可以轻而易举地穿女巫的衣服，完全不用担心被毒杀掉，所以狼人在首置位跳女巫的收益其实是很大的，因为女巫已经无法自证身份了！我对9号的女巫身份存疑，但我作为一个普普通通的村民，也不可能直接把9号打成狼人，我认为9号是女巫的概率也就一半一半吧！1号的发言虽然与我的观点不太一样，但也是有自己的逻辑输出的，我认为1号大概率是好人。2号完全是在跟风1号的发言内容，我认为2号可能是跟风划水的狼人牌。我等待着我的盖世预言家起跳吧！\\
Player 4: Good morning everyone, I'm your ordinary villager. Last night, players 3 and 8 both died, indicating the Witch used both potions on the first night. This makes it easy for werewolves to pretend to be the Witch without worrying about being poisoned, so there's actually a big benefit for werewolves to claim to be the Witch in the first position because the Witch can no longer prove their identity! I'm skeptical about player 9's Witch identity, but as a simple villager, I can't directly accuse 9 of being a werewolf. I'd say the probability of 9 being the Witch is about 50-50! Although player 1's statement doesn't quite align with my view, they did provide their own logical output, so I think player 1 is likely a good person. Player 2 is completely following player 1's statement content, I think 2 might be a werewolf card going with the flow. I'm waiting for my great Seer to step up! \\
5号玩家：前置位1号、2号和4号都表示自己是村民，我的底牌也是村民，说明前置位这三名玩家里面一定有狼人穿了村民的衣服。后面只有6号和7号了，如果只有一名玩家起跳预言家，我可能会考虑3号和8号里面有被女巫毒掉的预言家。我严厉谴责女巫不救人而毒人的做法，因为毒中好人的概率远比狼人高！\\
Player 5: The players in earlier positions - 1, 2, and 4 - all claim to be villagers, and my identity is also a villager, which means there must be a werewolf among these three players pretending to be a villager. There are only players 6 and 7 left, if only one player claims to be the Seer, I might consider that either player 3 or 8 was the Seer poisoned by the Witch. I strongly condemn the Witch's action of poisoning instead of saving, because the probability of poisoning a good person is much higher than a werewolf! \\ 
6号玩家：6号是村民。我前面已经有四张认村民的玩家了，其中必然有狼人。我认同5号玩家的观点，所以我认为5号是好人。后面只有7号一张牌了，我认为7号只能是预言家或狼人。我可能会考虑出局的玩家里面有被女巫毒掉的预言家，7号的发言如果不能打动我，那我可能会投7号。我也严厉谴责女巫不救人而毒人的做法，因为毒中好人的概率远比狼人高！
\\
Player 6: Player 6 is a villager. There are already four players claiming to be villagers before me, among which there must be a werewolf. I agree with player 5's view, so I think player 5 is a good person. There's only player 7 left, I think 7 can only be either the Seer or a werewolf. I might consider that the Seer was among the eliminated players and was poisoned by the Witch. If player 7's statement doesn't convince me, I might vote for 7. I also strongly condemn the Witch's action of poisoning instead of saving, because the probability of poisoning a good person is much higher than a werewolf! 
}
\end{tcolorbox}
\begin{tcolorbox}
{\ttfamily\small
7号玩家：大家早上好，我是你们的盖世预言家，5号查杀。查验5号就是想去看清5号的身份。先讲一下5号狼人的爆点，5号说“我可能会考虑3号和8号里面有被女巫毒掉的预言家”，人家9号女巫已经说了昨晚3号吃刀，8号被毒杀，而你5号的发言就是表示3号是自刀的狼人，8号是吃毒的好人，所以你是一张有视野的狼人牌！我认为3号和5号是两张狼人，今晚我要去查验6号的身份。我认为1号和4号都是有正常逻辑输出的好人，2号的发言过于划水。6号认为5号是好人，因此我认为6号大概率是狼人。今天好人全票出5号狼人，守卫去守护9号女巫，保证今晚平安夜。\\
Player 7: Good morning everyone, I'm your great Seer, player 5 is a werewolf. I checked player 5 because I wanted to clarify their identity. First, let me point out the giveaway for player 5 being a werewolf. Player 5 said "I might consider that either player 3 or 8 was the Seer poisoned by the Witch", but player 9, the Witch, already said that player 3 was killed by werewolves last night and player 8 was poisoned. So your statement, player 5, implies that player 3 was a self-stabbing werewolf and player 8 was a good person who was poisoned, which shows you're a werewolf card with inside information! I believe players 3 and 5 are two werewolves, and tonight I will check player 6's identity. I think players 1 and 4 are good people with normal logical output, while player 2's statement is too vague. Player 6 thinks player 5 is a good person, so I think player 6 is likely a werewolf. Today, all good people should vote for player 5, the werewolf, and the Guard should protect player 9, the Witch, to ensure a peaceful night tonight. \\\\
你目前是7号狼人。请综合角色设定、客观信息以及主观信息分析场上目前的局势并形成你本轮的笔记（注意客观信息一定为真实的，主观信息可能包含欺骗性的发言），要求对夜晚信息和玩家发言进行总结和分析，并输出投票原因和要投票出局的玩家，直接输出玩家编号数字；如果弃票，请输出“弃票”。请用关键字为“笔记”、“投票原因”和“投票玩家”的json格式输出。
\\
You are currently werewolf 7. Please analyze the current situation based on the role setting, objective information, and subjective information, and form your notes for this round (note that objective information is definitely true, while subjective information may contain deceptive statements). Summarize and analyze the night information and player statements, and output the reason for voting and the player to be voted out, directly output the player number; if abstaining, please output "abstain". Please output in JSON format with keywords "notes", "voting reason" and "voting player".
\\\\\\
\textbf{Response:}\\
\{"笔记": "夜晚信息：\\- 狼人，刀3号。\\- 3号和8号死亡。\\发言内容概括：\\9号玩家：\\- 跳女巫，昨晚3号被刀没有救，毒杀了8号。\\1号玩家：\\- 跳村民，认为9号是女巫。\\2号玩家：\\- 跳村民，认为9号是女巫。\\4号玩家：\\- 跳村民，认为9号可能是狼人，认为1号是好人，认为2号是狼人。\\5号玩家：\\- 跳村民，认为1号、2号和4号中有狼人，认为3号和8号中有被毒杀的预言家。\\6号玩家：\\- 跳村民，认为5号是好人，认为7号可能是预言家或狼人。\\- 可能会投7号出局。\\7号玩家：\\- 跳预言家，给5号发查杀，认为3号是狼人，8号是好人。\\- 要去查验6号的身份，认为1号和4号是好人，认为6号像狼人，要投5号出局，让守卫去守护9号女巫，保证今晚平安夜。",
}
\end{tcolorbox}
\begin{tcolorbox}
{\ttfamily\small
"投票原因": "作为狼人，7号（我）悍跳预言家给5号发查杀。此时我应该做的：\\- 投给5号玩家：通常情况下，作为预言家一定要优先投票给自己查验出来的狼人，否则将无法取得好人的信任。为了向好人证实我的“预言家”的身份，骗取好人的信任，我必须投给我的“查杀”5号。"投票玩家": "5"\}
\\
\{"notes": "Night information: \\
Werewolves killed player 3. \\
Players 3 and 8 died. \\ 
Summary of statements:  \\
Player 9: Claims to be the Witch, didn't save player 3 last night who was killed, poisoned player 8. \\
Player 1: Claims to be a villager, believes player 9 is the Witch. \\
Player 2: Claims to be a villager, believes player 9 is the Witch. \\
Player 4: Claims to be a villager, thinks player 9 might be a werewolf, believes player 1 is good, thinks player 2 is a werewolf.\\
Player 5:  Claims to be a villager, thinks there's a werewolf among players 1, 2, and 4, believes either player 3 or 8 was the Seer poisoned by the Witch. \\
Player 6: Claims to be a villager, thinks player 5 is good, believes player 7 might be the Seer or a werewolf. Might vote to eliminate player 7.\\
Player 7: Claims to be the Seer, accuses player 5 of being a werewolf, thinks player 3 is a werewolf and player 8 is good. Plans to check player 6's identity, believes players 1 and 4 are good, thinks player 6 seems like a werewolf, wants to vote out player 5, asks the Guard to protect player 9 (the Witch) to ensure a peaceful night tonight.",\\
"voting reason": "As a werewolf, player 7 (me) boldly claims to be the Seer and accuses player 5 of being a werewolf. What I should do now: \\
Vote for player 5: Usually, as the Seer, one must prioritize voting for the werewolf they've identified, otherwise they won't gain the trust of the good players. To prove my 'Seer' identity to the good players and gain their trust, I must vote for my 'werewolf accusation', player 5.", "voting player": "5" \}
}
\end{tcolorbox}

\textbf{Role Prediction}
\begin{tcolorbox}
{\ttfamily\small
\textbf{Prompt:}\\
在本场游戏中，你目前已知以下信息：\\
In this game, you currently have the following information: \\

1. 角色设定：你是1号玩家。\\
你的身份是：狼人。\\
你需要其他狼人合作，每晚选择一个村民猎杀，你的目标是隐藏你的身份，并误导其他玩家，直至狼人获得游戏的胜利。\\
1. Role setup: You are Player 1.\\ 
Your identity is: Werewolf.\\ 
You need to cooperate with other werewolves, choose a villager to hunt each night, and your goal is to hide your identity and mislead other players until the werewolves win the game. \\\\
2. 客观信息：\\- 游戏进程：目前游戏进行到第1轮。
\\- 狼人为：1,6,7号玩家。
\\- 当前存活的玩家有：1号，2号，4号，5号，6号，7号，9号，
\\- 夜晚信息：第1轮8,3号玩家死亡。
\\- 投票情况：暂无 
}
\end{tcolorbox}
\begin{tcolorbox}
{\ttfamily\small
2. Objective information: \\
Game progress: The game is currently in Round 1. \\
The werewolves are: Players 1, 6, and 7. \\
Currently surviving players are: Players 1, 2, 4, 5, 6, 7, and 9.
\\
Night information: In Round 1, Players 8 and 3 died. \\
Voting situation: None yet \\\\
3. 主观信息：\\第1轮所有玩家的发言为：\\
3. Subjective information: \\ 
Round 1 speeches from all players: \\
9号玩家：昨晚3号被狼人杀害，我没救，毒杀了8号。9号是女巫，场上的守卫不要报身份，自己藏好，今晚守护我，我必去救外置位的牌，今天谁跳守卫我必投票给他。对话全场好人，如果后置位有狼人悍跳女巫想将我抗推出局，我希望好人能够相信我是真女巫。如果不信我是真女巫也没关系，悍跳女巫的狼人为了证明自己是真女巫，在女巫只剩一瓶解药的情况下，他也要空刀保证一轮平安夜，否则第二天他可以直接出局。我认为狼人悍跳预言家总会有漏洞，我会根据发言选择站边。
\\
Player 9: Last night, Player 3 was killed by werewolves, I didn't save him, but I poisoned Player 8. I'm the Witch. The Guard shouldn't reveal their identity, stay hidden, and protect me tonight. I will definitely save an outer position player tonight. If anyone claims to be the Guard today, I will vote for them. I'm addressing all the good players: if there's a werewolf in the later positions who's falsely claiming to be the Witch and trying to get me voted out, I hope the good players will believe I'm the real Witch. If you don't believe I'm the real Witch, that's fine too. A werewolf falsely claiming to be the Witch, to prove they're the real Witch, will have to ensure a peaceful night in the situation where the Witch only has one antidote left, otherwise they can be directly voted out on the second day. I believe werewolves falsely claiming to be the Seer will always have flaws, and I will choose sides based on the speeches. \\
1号玩家：我是村民，夜间3号、8号出局，或许正如9号玩家所言，9号是女巫没有救3号反而毒了8号。对于已出局的3号和8号，我不作任何评价，他们的身份是什么与我无关。我认为9号发言很诚恳，符合女巫的视角，且狼人第一个发言就悍跳女巫貌似没有太大的收益，因此我认为9号大概率是真女巫。我等待预言家报验人吧，我这边就先过掉。\\
Player 1: I'm a villager. During the night, Players 3 and 8 were eliminated. Perhaps as Player 9 said, Player 9 is the Witch who didn't save Player 3 but poisoned Player 8 instead. Regarding the eliminated Players 3 and 8, I won't make any comments, their identities are irrelevant to me. I think Player 9's speech was sincere and aligns with the Witch's perspective. Also, it doesn't seem very beneficial for a werewolf to falsely claim to be the Witch in the first speech, so I believe Player 9 is likely the real Witch. I'll wait for the Seer to report their verification. I'll pass for now. \\
2号玩家：2号是村民，我相信9号是女巫，因为狼人第一个发言就悍跳女巫没有太大的收益，因此我认为9号大概率是真女巫。1号的发言我看不出是什么身份，等待跳预言家的玩家发言吧，我再去站边。\\
Player 2: Player 2 is a villager. I believe Player 9 is the Witch because it's not very beneficial for a werewolf to falsely claim to be the Witch in the first speech, so I think Player 9 is likely the real Witch. I can't tell what identity Player 1 is from their speech. Let's wait for players claiming to be the Seer to speak, then I'll choose sides. \\
4号玩家：大家早上好，我是你们的盖世村民。昨晚3号、8号双死，说明女巫在第一晚就开天毒了，这使得狼人可以轻而易举地穿女巫的衣服，完全不用担心被毒杀掉，所以狼人在首置位跳女巫的收益其实是很大的，因为女巫已经无法自证身份了！我对9号的女巫身份存疑，但我作为一个普普通通的村民，也不可能直接把9号打成狼人，我认为9号是女巫的概率也就一半一半吧！1号的发言虽然与我的观点不太一样，但也是有自己的逻辑输出的，我认为1号大概率是好人。2号完全是在跟风1号的发言内容，我认为2号可能是跟风划水的狼人牌。我等待着我的盖世预言家起跳吧！
}
\end{tcolorbox}

\begin{tcolorbox}
{\ttfamily\small
Player 4: Good morning everyone, I'm your ordinary villager. Last night, Players 3 and 8 both died, which means the Witch used both potions on the first night. This makes it easy for werewolves to falsely claim to be the Witch without worrying about being poisoned, so there's actually a big benefit for werewolves to claim to be the Witch in the first position because the Witch can no longer prove their identity! I'm skeptical about Player 9's Witch identity, but as a simple villager, I can't directly accuse Player 9 of being a werewolf. I'd say the probability of Player 9 being the Witch is about 50-50! Although Player 1's speech doesn't quite align with my view, they did provide their own logical output, so I think Player 1 is likely a good person. Player 2 is completely following Player 1's speech content, so I think Player 2 might be a werewolf just going with the flow. I'm waiting for my great Seer to step up! \\
5号玩家：前置位1号、2号和4号都表示自己是村民，我的底牌也是村民，说明前置位这三名玩家里面一定有狼人穿了村民的衣服。后面只有6号和7号了，如果只有一名玩家起跳预言家，我可能会考虑3号和8号里面有被女巫毒掉的预言家。我严厉谴责女巫不救人而毒人的做法，因为毒中好人的概率远比狼人高！\\
Player 5: The front positions, Players 1, 2, and 4, all claim to be villagers. My identity is also a villager, which means there must be a werewolf among these three front position players pretending to be a villager. There are only Players 6 and 7 left. If only one player claims to be the Seer, I might consider that either Player 3 or 8 was the Seer poisoned by the Witch. I strongly condemn the Witch's action of poisoning instead of saving, because the probability of poisoning a good person is much higher than poisoning a werewolf!\\
6号玩家：6号是村民。我前面已经有四张认村民的玩家了，其中必然有狼人。我认同5号玩家的观点，所以我认为5号是好人。后面只有7号一张牌了，我认为7号只能是预言家或狼人。我可能会考虑出局的玩家里面有被女巫毒掉的预言家，7号的发言如果不能打动我，那我可能会投7号。我也严厉谴责女巫不救人而毒人的做法，因为毒中好人的概率远比狼人高！\\
Player 6: Player 6 is a villager. There are already four players claiming to be villagers before me, among which there must be a werewolf. I agree with Player 5's view, so I think Player 5 is a good person. There's only Player 7 left, so I think Player 7 can only be either the Seer or a werewolf. I might consider that the Seer was among the eliminated players and was poisoned by the Witch. If Player 7's speech doesn't convince me, I might vote for Player 7. I also strongly condemn the Witch's action of poisoning instead of saving, because the probability of poisoning a good person is much higher than poisoning a werewolf! 
\\
7号玩家：大家早上好，我是你们的盖世预言家，5号查杀。查验5号就是想去看清5号的身份。先讲一下5号狼人的爆点，5号说“我可能会考虑3号和8号里面有被女巫毒掉的预言家”，人家9号女巫已经说了昨晚3号吃刀，8号被毒杀，而你5号的发言就是表示3号是自刀的狼人，8号是吃毒的好人，所以你是一张有视野的狼人牌！我认为3号和5号是两张狼人，今晚我要去查验6号的身份。我认为1号和4号都是有正常逻辑输出的好人，2号的发言过于划水。6号认为5号是好人，因此我认为6号大概率是狼人。今天好人全票出5号狼人，守卫去守护9号女巫，保证今晚平安夜。
\\
Player 7: Good morning everyone, I'm your great Seer, and Player 5 is my kill check. I checked Player 5 because I wanted to clarify their identity. First, let me point out Player 5's werewolf tells. Player 5 said, "I might consider that either Player 3 or 8 was the Seer poisoned by the Witch," but Player 9, the Witch, already said that Player 3 was killed by werewolves last night and Player 8 was poisoned. So Player 5's speech implies that Player 3 was a self-stabbing werewolf and Player 8 was a good person who was poisoned, which shows that Player 5 is a werewolf with inside information!\\
I believe Players 3 and 5 are both werewolves. Tonight, I will check Player 6's identity. I think Players 1 and 4 are both good people with normal logical output, while Player 2's speech is too passive. Player 6 thinks Player 5 is a good person, so I think Player 6 is likely a werewolf. Today, all good people should vote for Player 5, the werewolf, and the Guard should protect Player 9, the Witch, to ensure a peaceful night tonight.
}
\end{tcolorbox}

\begin{tcolorbox}
{\ttfamily\small
你目前是1号狼人。请综合角色设定、客观信息以及主观信息（注意客观信息一定为真实的，主观信息可能包含欺骗性的发言）预测所有玩家的身份标签。请用关键字为“N号玩家”的json格式输出。\\
You are currently Werewolf 1. Please predict the identity labels for all players based on the role setup, objective information, and subjective information (note that objective information is always true, while subjective information may contain deceptive speeches). Please output in JSON format using the keyword "Player N" for each player. \\
\\
\textbf{Response:}\\
\{"1号玩家": "狼人", "2号玩家": "普通村民", "3号玩家": "预言家", "4号玩家": "普通村民", "5号玩家": "普通村民", "6号玩家": "狼人", "7号玩家": "狼人", "8号玩家": "守卫", "9号玩家": "女巫"\} \\
\{"Player 1": "Werewolf", "Player 2": "Villager", "Player 3": "Seer", "Player 4": "Villager", "Player 5": "Villager", "Player 6": "Werewolf", "Player 7": "Werewolf", "Player 8": "Guard", "Player 9": "Witch"\}
}
\end{tcolorbox}
\end{CJK}

\section{MaKTO Details}
\label{sec:app:makto_details}

\paragraph{Loss Function of Kahneman-Tversky Optimization }
We assign desirable and undesirable behaviors manually and use KTO to update the model. 
Given the prompt-response from dataset \((x, y) \in D\), KTO uses the following loss to optimize the policy \(\pi_\theta\):

\[r_\theta(x,y) = \log \frac{\pi_\theta(y|x)}{\pi_{ref}(y|x)} \]

\[z_0 = \mathbb{E}_{(x,y)\sim D}[KL(\pi_\theta(y|x) \| \pi_{ref}(y|x)) ]  \]

\[
v(x,y) = 
\left\{
\begin{array}{ll}
\lambda_D\sigma(\beta(r_\theta(x,y)-z_0)), & \text{if } y \sim y_{desirable}|x \\
\lambda_U\sigma(\beta(z_0-r_\theta(x,y))), & \text{if } y \sim y_{undesirable}|x \\
\end{array}
\right.
\]

Therefore, 

\[L(\pi_\theta;D) = \mathbb{E}_{(x,y)\sim D}[\lambda_y - v(x,y)]\] 

Here, \( \lambda_D \) and \( \lambda_U \) are hyperparameters for the desirable and undesirable losses, respectively. 
The parameter \( \lambda_y \) represents \( \lambda_D \) when \( y \) is desirable and \( \lambda_U \) when \( y \) is undesirable.

\paragraph{Criteria of Stepwise Preference Data Selection}
The selection methods for the perference data were briefly described in Section~\ref{sec:method_kto}, and here, Table~\ref{tab:kto_selection} shows the detailed selection criteria.
\begin{table}[htbp]
\centering
\footnotesize
\resizebox{1\linewidth}{!}{
    \begin{tabular}{p{1.3cm}|p{1cm}|p{5cm}|p{5cm}}
    
    \toprule
    \textbf{Selection Method} & \textbf{Game Stage} & \makecell[c]{\textbf{Desirable}} & \makecell[c]{\textbf{Unacceptable}} \\
    \midrule
    
    \multirow{12}{*}[-2em]{\makecell[c]{\textbf{Heuristic}}} & 
    \multirow{6}{*}[-1em]{\makecell[c]{Night\\Action}} & 
    \begin{itemize}[noitemsep, topsep=0pt, parsep=0pt, partopsep=0pt, leftmargin=*]
        \item Werewolves targeting special roles from day 2
        \item Seer identifying a werewolf
        \item Witch saving someone on night 1
        \item Witch poisoning a werewolf from day 2
        \item Guard protecting special roles from day 2
        \item Hunter correctly eliminating a werewolf
    \end{itemize} & 
    \begin{itemize}[noitemsep, topsep=0pt, parsep=0pt, partopsep=0pt, leftmargin=*]
        \item Werewolves not attacking anyone
        \item Witch not saving anyone on the first night
        \item Witch poisoning a villager from day 2
        \item Guard protecting a werewolf
        \item Hunter eliminating a special role
    \end{itemize} \\
    \cmidrule{2-4}
    
    & \multirow{6}{*}[-1em]{\makecell[c]{Vote}} & 
    \begin{itemize}[noitemsep, topsep=0pt, parsep=0pt, partopsep=0pt, leftmargin=*]
        \item Villagers voting for and successfully eliminating a werewolf
        \item Special roles voting for a werewolf
    \end{itemize} & 
    \begin{itemize}[noitemsep, topsep=0pt, parsep=0pt, partopsep=0pt, leftmargin=*]
        \item Villagers voting for and eliminating another villager
        \item Abstaining from voting
        \item Not voting with the true Seer (splitting votes with the Seer)
        \item If no Seer is present, not voting with the majority of villagers (splitting votes with most villagers)
    \end{itemize} \\
    \midrule
    
    \multirow{6}{*}[-2em]{\makecell[c]{\textbf{Staged}\\\textbf{voting}}} & 
    \multirow{6}{*}[-1em]{\makecell[c]{Speech}} & 
    \begin{itemize}[noitemsep, topsep=0pt, parsep=0pt, partopsep=0pt, leftmargin=*]
        \item Werewolf speaking without being voted out
        \item Villager receiving no votes
        \item Seer receiving less than one villager vote
    \end{itemize} & 
    \begin{itemize}[noitemsep, topsep=0pt, parsep=0pt, partopsep=0pt, leftmargin=*]
        \item Werewolf speaking and being voted out
        \item Werewolf speaking and receiving more than half of villager votes
        \item All villagers speaking and being voted out
        \item Witch speaking and receiving werewolf votes \& more than two villager votes
        \item Seer speaking and receiving more than half of villager votes
    \end{itemize} \\
    \midrule
    
    \multirow{2}{*}[-1em]{\makecell[c]{\textbf{Verifier-}\\\textbf{based}}} & 
    \multirow{2}{*}[-1em]{\makecell[c]{Speech}} & 
    \begin{itemize}[noitemsep, topsep=0pt, parsep=0pt, partopsep=0pt, leftmargin=*]
        \item Speech with no conflict with the observable fact in the gameplay
    \end{itemize} &
    \begin{itemize}[noitemsep, topsep=0pt, parsep=0pt, partopsep=0pt, leftmargin=*]
        \item Speech that conflicts with the fact.
    \end{itemize} \\
    \bottomrule
    \end{tabular}
}

\caption{Selection methods for Werewolf game actions.}
\label{tab:kto_selection}
\end{table}

\section{Training Details}
\label{sec:app:train}
% \paragraph{Base model}
Given that our collected expert training data is based on Chinese, and considering a stronger understanding of the Chinese context, we choose \texttt{Qwen2.5-14b-instruct} and \texttt{Qwen2.5-72b-instruct}~\cite{yang2024qwen2} as the base models for training. 

% \paragraph{SFT}
The SFT dataset comprises 25k samples, including 380 samples of fundamental game comprehension data with terminology explanations, 372 Q\&As on advanced gaming techniques, 12k annotated authentic gaming behavior data, and 12k general SFT corpus. 
We employed DeepSpeed ZeRO-3 optimization with a learning rate of $1e-6$, a warm-up ratio of $0.05$, and trained for 3 epochs.

% \paragraph{Multi-agent KTO}
For the Multi-agent KTO phase, we collected 20k preference data entries from the Seer-Witch-Guard games, consisting of 12k desirable and 8k unacceptable samples. The model pool includes \texttt{GPT-4o\_mini}, \texttt{GPT-4o}, fine-tuned \texttt{Qwen2.5-14b-instruct}, fine-tuned \texttt{Llama-3.1-8B-Instruct}, and fine-tuned \texttt{Qwen2.5-72b-instruct}. 
We set the KTO hyper-parameters with  \( \lambda_D =0.7\) and \( \lambda_U=1.0 \). 
The training utilized DeepSpeed ZeRO-3 optimization, with a learning rate of $1e-6$, a batch size of 2 per device, 150 warmup steps, and train for 20 epochs.
The 14B models are trained using 8 A100 GPUs and the 72B models used 32 A100 GPUs.

\section{Details of the Behavior Analysis Experiments}
\label{sec:app:behavior}

Here, we list detailed explanations of metrics used in Behavior Analysis~(Sec.~\ref{sec:result:behavior_model}), as well as specific performance of SFT and MaKTO models (against \texttt{GPT-4o} opponents) on these metrics:
These are detailed explanations of metrics used in Behavior Analysis, as well as specific performance of SFT and MaKTO models on these metrics:

% \begin{itemize}
\noindent\textbf{When Models Play as Villagers:}

    \begin{itemize}
    \item \textbf{Vote Acc.}: measures how accurately villagers vote for actual werewolves. This is a key indicator of villagers' ability to identify threats. Higher values indicate better overall performance.
    
    \item \textbf{Abstention}: the frequency of villagers choosing not to vote. This reflects their decision-making confidence in the game. \underline{Lower} values in this category indicate better performance, as it shows more decisive action-taking.
    
    \item \textbf{Werewolf Check}: specific to the Seer role and measures their success rate in identifying werewolves during the second night. Higher values demonstrate better deductive reasoning based on the first day's interactions.
    
    \item \textbf{Save @ Night 1}: applies specifically to the Witch role, measuring the rate of successfully saving players on the first night. Since the Witch cannot determine whether werewolves employed a self-attacking strategy during the first night, saving a targeted player is considered the safest to protect potential crucial roles (such as the Seer). Higher values show better strategic use of the rescue potion, suggesting the witch can make conservative and protective decisions early in the game.
    
    \item \textbf{Correct Poison}: relates to the Witch's ability to successfully poison actual werewolves. Higher values indicate better accuracy in threat identification and strategic decision-making.
    
    \item \textbf{Mispoison}: tracks the Witch's rate of accidentally poisoning fellow villagers. \underline{Lower} values in this metric indicate better judgment and decision-making abilities.
    
    \item \textbf{Protect God}: focuses on the Guard's success rate in protecting special role villagers. Successfully protecting special-role villagers gives the villager team a better chance of winning. Higher values indicate that the Guard accurately identifies teammates, especially those with special roles, and correctly uses their protection skill. 
    
    \item \textbf{Misprotect}: measures the Guard's rate of wrongly protecting werewolves. \underline{Lower} values indicate better accuracy in distinguishing between villagers and werewolves, meaning fewer instances where the Guard mistakenly protects a werewolf player.
    \end{itemize}

Table~\ref{tab:kto_fine_grain_villager} shows a detailed comparison of the fine-grained values of the metrics between the MaKTO and SFT models.

\begin{table*}[th]
    \small
    \centering
    \begin{tabular}{l|rr|r|rrr|rr}
    \toprule
    & \multicolumn{2}{|c}{\textbf{as Villagers}} & \multicolumn{1}{|c}{\textbf{as Seer}} & \multicolumn{3}{|c}{\textbf{as Witch}} & \multicolumn{2}{|c}{\textbf{as Guard}} \\
    \midrule

    & \shortstack{Vote\\Acc.($\uparrow$)}
    & \shortstack{Abstent-\\ion($\downarrow$)}
    & \shortstack{Werewolf\\Check($\uparrow$)} 
    & \shortstack{Save @\\Night 1($\uparrow$)} 
    & \shortstack{Correct\\Poison($\uparrow$)} 
    & \shortstack{Mis-\\poison($\downarrow$)}
    & \shortstack{Protect\\God($\uparrow$)} 
    & \shortstack{Mis-\\protect($\downarrow$)} \\
     \midrule
     
    \textbf{SFT-72b} & 66.3\% & 6.7\% & 60.0\% & 58.0\% & 48.0\% & 38.0\% & 34.1\% & 9.5\% \\
    \textbf{MaKTO-72b} & 72.7\% & 3.2\% & 75.7\% & 100.0\% & 72.0\% & 26.0\% & 37.6\% & 2.3\% \\
    \bottomrule
    \end{tabular}
    \caption{Fine-grained metrics of SFT-72b and MaKTO-72b model when act as villagers against \texttt{GPT-4o} werewolf.}
    \label{tab:kto_fine_grain_villager}
\end{table*}

\noindent\textbf{When Models Play as Werewolves} (measured through \texttt{GPT-4o} opponents' behavior):
    \begin{itemize}
    
    \item \textbf{Abstention}: the rate of opponents' voting abstention. A higher rate indicates that werewolf models' deceptive tactics were successful in creating enough confusion to prevent \texttt{GPT-4o} villagers from making voting decisions.
        
    \item \textbf{Seer Check}: the rate of successful werewolf identification by opponent Seer. \underline{Lower} values demonstrate that the werewolves' deceptive speeches were more effective, causing the Seer to misdirect their investigations toward innocent players rather than real werewolves.
    
    \item \textbf{Mispoison}: the rate of opponent Witch poisoning innocent villagers. 
    Higher values indicate that werewolf models' misdirections were more effective in making \texttt{GPT-4o} Witch suspicious of her own team.
        
    \item \textbf{Misprotect} the rate of opponent Guard protecting werewolves. Higher values indicate more effective deception and manipulation strategies by the werewolf team,  resulting in the Guard mistakenly protecting werewolves.
    
    \end{itemize}
Table~\ref{tab:kto_fine_grain_wolf} shows the specific values of the above metrics above when the model is a werewolf. The MaKTO werewolf is better at camouflaging than the baseline SFT model.
\begin{table*}[ht]
    \small
    \centering

    \begin{tabular}{l|rrrr}
        \toprule
        \textbf{Opponent's $\rightarrow$} 
        & \%\shortstack{\textbf{Villagers}\\\textbf{Abstention($\uparrow$)}}
        % & \%\textbf{Villagers' Abstention}($\uparrow$) 
        & \textbf{\% Seer Check ($\downarrow$)}
        & \%\shortstack{\textbf{Witch}\\\textbf{Mispoison($\uparrow$)}}
        % & \%\textbf{Witch's} Mispoison ($\uparrow$) 
        & \%\shortstack{\textbf{Guard}\\\textbf{Misprotect($\uparrow$)}} \\
        % & \%\textbf{Guard's} Protection ($\uparrow$)\\
        \midrule
        \textbf{SFT-72b} & 2.2\%	& 60.4\% & 32.0\% & 33.6\% \\
        \textbf{MaKTO-72b}	& 4.6\%	 & 57.4\% & 48.0\%	& 34.7\% \\
        \bottomrule
    \end{tabular}
    \caption{Fine-grained metrics of SFT-72b and KTO-72b when acting as werewolves against \texttt{GPT-4o} villagers. Note that Werewolf agent performance is reflected by opponent villager behavior. Lower opponent performance indicates a stronger werewolf model.}
    \label{tab:kto_fine_grain_wolf}
\end{table*}

\section{Ablation: Is MaKTO more effective than SFT only on desirable data?}
\label{sec:app:sft_vs_kto}
From the annotated data, we can also select desirable data using heuristic-based and staged voting-based methods. 
Would SFT based solely on this desirable data perform better? We conduct a win rate comparison experiment as shown in Table~\ref{tab:kto_vs_sft}. 
The experimental results show that MaKTO-72B achieves a remarkable 0.593 average win rate, while SFT with desirable data falls short at 0.483, actually showing a slight decrease of -0.02 compared to the baseline. In direct competition, MaKTO-72B maintains an edge over SFT with desirable data, securing a 0.53 win rate. This may be due to the reduced total amount of data when selecting only desirable data. MaKTO's advantages in strategic depth and adaptability surpass what can be achieved through SFT on desirable data alone.

\begin{table}[h]
    \small
    \centering
    \begin{tabular}{l|rr|rr|l}
    \toprule
     & \textbf{GPT-4o} & \textbf{SFT-72b} & \textbf{MaKTO-72b}
     & \shortstack{\textbf{SFT w/} \\\textbf{desirable}} 
     & \textbf{Avg. Win Rate}\\
     % & SFT w/ desirable & avg \\
    \midrule
    \textbf{GPT-4o} & {\color[HTML]{FF0000} 0.50} & 0.40 & 0.35 & \textbf{0.52} & 0.423 \\
    \textbf{SFT-72b} & \textbf{0.60} & {\color[HTML]{FF0000} 0.50} & 0.40 & 0.50 & 0.500 \\
    \midrule
    \textbf{MaKTO}-72b & \textbf{0.65} & \textbf{0.60} & {\color[HTML]{FF0000} 0.50} & \textbf{0.53} & 0.593 $(+0.09)$ \\
    \textbf{SFT w/ desirable} & 0.48 & 0.50 & 0.47 & {\color[HTML]{FF0000} 0.50} & 0.483 $(-0.02)$\\
    \bottomrule
    \end{tabular}
    \caption{Average win rate for MaKTO and SFT model trained only using desirable data in 9-player Seer-Witch-Guard game. A win rate above 0.5 (in \textbf{bold}) indicates that the model in the row significantly outperforms the model in the column. All models are in 72b size.}
    \label{tab:kto_vs_sft}
\end{table}

\clearpage
\section{Case Study}
\label{sec:app:case}
\section{Cases}
\label{app:case}
% \paragraph{Human-AI test.}

The following is a case of a human-AI head-to-head competition with MaKTO-72b as villagers and human players as werewolves.
The gameplay is roughly as follows.
\underline{Night 1}: Witch saves werewolf. \underline{Day 1}: Real Seer voted out after werewolf's (human) convincing fake claim. \underline{Night 2}: Guard protects Witch. \underline{Day 2}: Villager team (MaKTO) realize the mistake day and eliminates fake Seer. \underline{Night 2}: Witch successfully poisons a werewolf. \underline{Day 3}: Final werewolf eliminated after failed Guard claim.

In this case, we find that the MaKTO models played their roles accurately and were able to spot some humorous statements in the discussion period and make appropriate strategic decisions.
\textcolor{red}{Remarkable speeches and actions are marked in red.}
They demonstrated:
\textit{1)} logical analysis and deduction --- a villager (Player 9) keenly spotted flaws fake Seer's lies on Day 1.
\textit{2)} Strategic adaptation --- despite misjudgment on Day 1 that eliminated the real Seer, they quickly adjusted their strategy.
\textit{3)} Team collaboration --- MaKTO models showed excellent coordination, especially during Day 2 voting.
\textit{4)} Strategic use of role abilities --- Guard accurately protected the key Witch role, and the Witch successfully poisoned a werewolf.

\begin{CJK}{UTF8}{gkai}

\begin{tcolorbox}
{\ttfamily\small
\textbf{Role Assignments:}\\
* Player 1 (MaKTO-72B): 守卫 Guard.\\
* Player 2 (Human):\quad\qquad狼人 Werewolf.\\
* Player 3 (Human):\quad\qquad狼人 Werewolf.\\
* Player 4 (MaKTO-72B): 预言家 Seer.\\
* Player 5 (MaKTO-72B): 普通村民 Simple villager.\\
* Player 6 (MaKTO-72B): 普通村民 Simple villager.\\
* Player 7 (Human):\quad\qquad狼人 Werewolf.\\
* Player 8 (MaKTO-72B): 女巫 Witch.\\
* Player 9 (MaKTO-72B): 普通村民 Simple villager.\\
\\
\textbf{Night 1:}\\
* \textbf{Werewolves}: \\
\vspace{2pt} Player 2, Player 3 and Player 7 chose to kill Player 7.\\
* \textbf{Seer}: Player 4 checked Player 8 -- Not a Werewolf.\\
* \textbf{Witch}: \\
\vspace{2pt} Player 8 \textcolor{red}{used antidotes to save Player 7}; did not use poison. \\
* \textbf{Guard}: \textcolor{red}{Player 1 chose to protect None.}\\
\\
\textbf{Day 1 announcement:}\\
No one died last night.\\
Start discussion from Player 4.\\
\\
\textbf{Day 1 discussion:}\\
\textbf{* Player 4 (MaKTO-72B, Seer)}: \\
我是预言家，昨晚查验9号是好人，前置位第一个发言没什么多余的信息了，本轮归票后置位悍跳狼。\\
I am the Seer. Last night, I verified that Player 9 is a good person. As the first speaker in the front positions, I don't have much additional information to share. This round, I'm voting for the Werewolf who is boldly claiming to be the Seer in the back positions.\\
\textbf{* Player 5 (MaKTO-72B, Simple villager)}：\\
\textcolor{red}{吾乃一介村民，不才，昨晚睡得有点死，居然不知道昨晚是哪个盖世守卫和女巫大人联手维护了村庄和平，当然也不排除是狼人自刀的可能性}。不过这些都不重要了，重要的是今天4号起跳预言家给9号发金水，发言简短，没有视角，可能是真预言家也可能是夜间没有安排工作的狼人。\textcolor{red}{我需要根据9号的发言来选择站边}，如果9号的发言取得我的好感，我可能会站边4号，否则，我将会考虑4号是悍跳狼人。
}
\end{tcolorbox}

\begin{tcolorbox}
{\ttfamily\small 
\textcolor{red}{I am but a humble villager. I must admit, I slept rather soundly last night and was unaware of which legendary guard and esteemed witch joined forces to maintain peace in our village. Of course, we can't rule out the possibility that the werewolves chose not to attack anyone.} However, these matters are no longer of great importance. What's crucial is that today, Player 4 has claimed to be the Seer and declared Player 9 as "golden"~(confirmed innocent). Their statement was brief and lacked perspective, suggesting they could either be the true Seer or a werewolf who had no assigned task during the night. \textcolor{red}{I need to base my decision on Player9's upcoming statement}. If Player 9's words resonate well with me, I might side with Player 4. Otherwise, I will consider the possibility that Player 4 is a werewolf boldly impersonating the Seer.
\\
\textbf{* Player 7 (MaKTO-72B, Simple villager)}: \\
我是一个村民，\textcolor{red}{4号跳预言家发言有些简短，我暂时先不选择站边}。我认为4号可能是预言家或狼人，5号可能是村民或狼人。先听一下后置位对跳预言家的玩家发言以及9号的发言吧，我会根据发言选择站边。\\
I am a Villager. \textcolor{red}{Player 4's claim to be the Seer was rather brief, so I'm not choosing sides yet}. I think Player 4 could be either the Seer or a werewolf, and Player 5 could be either a villager or a werewolf. Let's first listen to the statements from players in the back positions who might also claim to be the Seer, as well as Player 9's statement. I will choose sides based on these upcoming statements.\\
\textbf{* Player 7 (Human, Werewolf)}: \\
我是预言家，昨夜查验8号玩家是金水，查验的心路历程就是从身边查验，我如果是悍跳的狼人，给后置位发查杀显然力度更大。因为本局游戏有守卫在场，如果昨夜是女巫使用了解药，女巫和守卫可以先不要跳出来表明身份，守卫今夜来守护我，女巫可以选择心中的狼人进行毒杀。如果女巫昨夜没有使用解药，女巫和守卫可以站出来排一下坑，这样的话守卫晚上守护女巫，女巫解救被刀杀的玩家，或者毒杀像狼人的玩家。\\
I am the Seer. Last night, I verified that Player 8 is "golden" (confirmed innocent). My verification process was to check those near me. If I were a werewolf impersonating the Seer, it would have been more impactful to accuse a player in the back positions of being a werewolf. Since there's a Guard in this game, if the Witch used her healing potion last night, both the Witch and the Guard should refrain from revealing their identities for now. The Guard should protect me tonight, and the Witch can choose to poison a player she suspects to be a werewolf. If the Witch didn't use her healing potion last night, both the Witch and the Guard can come forward to clarify their positions. In this case, the Guard can protect the Witch at night, while the Witch can either save a player who's been attacked by werewolves or poison a player who seems suspicious.
\\
\textbf{* Player 8 (MaKTO-72B, Witch)}: \\
我是女巫，已使用解药解救7号，目前没有使用毒药。我认为前置位4号发言过于简短，没有考虑到女巫和守卫在场，没有安排守卫和女巫的工作，所以4号像一只悍跳狼人。而7号发言中规中矩，考虑到了场上有守卫和女巫，对女巫和守卫的工作也进行了安排，所以我认为7号像预言家。如果后置位有发言更好的预言家，我可能会考虑更换站边。\\
I am the Witch. I have already used my healing potion to save Player 7, but I haven't used my poison yet. I think Player 4's statement in the front positions was too brief and didn't consider the presence of the Witch and the Guard. Player 4 also didn't provide any instructions for the Witch and Guard's actions, so they seem like a werewolf boldly impersonating the Seer. On the other hand, Player 7's statement was appropriate and took into account the presence of the Guard and the Witch on the field. Player 7 also gave instructions for the Witch and Guard's actions, so I believe Player 7 is more likely to be the real Seer. If a player in the back positions makes a better claim as the Seer, I might reconsider my support. 
}
\end{tcolorbox}

\begin{tcolorbox}
{\ttfamily\small 
\textbf{* Player 9 (MaKTO-72B, Simple villager)}: \\
我是一张村民牌，我选择站边4号。\textcolor{red}{7号的爆点在于他说“我如果是悍跳的狼人，给后置位发查杀显然力度更大”，这是狼人视角的发言，真预言家不可能怀疑自己是狼人，也不可能去假设自己是悍跳狼，这是只有狼人才会去思考的逻辑。所以我认为7号大概率是悍跳狼，而4号大概率是真预言家。}8号起跳女巫，目前没有人对跳女巫，我认为8号大概率是真女巫。4号发言过于简短，我暂时无法判断其真实身份，他可能是预言家也有可能是狼人。本轮如果后面没有起跳预言家的玩家，我会投票给7号。\\
I am a Villager. I choose to side with Player 4. \textcolor{red}{Player 7's tell is in their statement, "If I were a werewolf impersonating the Seer, it would have been more impactful to accuse a player in the back positions of being a werewolf." This statement comes from a werewolf's perspective. A real Seer would never suspect themselves of being a werewolf, nor would they hypothesize about being a werewolf impersonating the Seer. This kind of logic is something only a werewolf would consider. Therefore, I believe Player 7 is most likely a werewolf boldly claiming to be the Seer, while Player 4 is probably the real Seer.}
Player 8 has claimed to be the Witch, and so far, no one else has contested this role. I believe Player 8 is likely the real Witch. Player 4's statement was too brief, so I can't determine their true identity yet - they could be either the Seer or a werewolf.
This round, if no other player claims to be the Seer later, I will vote for Player 7.
\\
\textbf{* Player 1 (MaKTO-72B, Guard)}: \\
我是守卫，昨晚空守。我认为7号是预言家，4号是悍跳狼。前置位4号和7号对跳预言家，4号给后置位的9号发金水，7号给身旁的8号发金水。4号的查验力度看似比7号的更大，但4号作为第一个发言的玩家，言简意赅也说得过去，所以单从两位跳预言家的玩家的发言去分析，难以分辨出谁是真预言家。但是9号的发言站边4号，攻击7号的理由并不成立，9号纯粹是因为只言片语去攻击7号是狼，我认为9号更像是知道7号是预言家，从而去强行攻击7号的狼人。相反8号接到7号的金水后跳女巫，发言真诚地分析两位跳预言家的玩家的优劣，并没有直接站边给自己的银水7号，我认为8号更像是真女巫。如果后置位没有跳女巫的玩家，我将会站边7号，投票给4号。\\
I am the Guard. Last night, I didn't protect anyone. I believe Player 7 is the Seer, and Player 4 is a werewolf boldly impersonating the Seer. In the front positions, Players 4 and 7 both claimed to be the Seer. Player 4 declared Player 9 in the back positions as "golden" (confirmed innocent), while Player 7 declared the adjacent Player 8 as "golden".
Player 4's verification seems more impactful than Player 7's at first glance, but since Player 4 was the first to speak, their brevity is understandable. So, analyzing just the statements of these two Seer claimants, it's difficult to determine who the real Seer is.
However, Player 9's statement supporting Player 4 and attacking Player 7 is not well-founded. Player 9 is attacking Player 7 as a werewolf based on mere fragments of speech, which makes me think Player 9 is more likely a werewolf who knows Player 7 is the real Seer and is forcefully attacking them.
In contrast, Player 8, after being declared "golden" by Player 7, claimed to be the Witch and sincerely analyzed the pros and cons of both Seer claimants without immediately siding with Player 7, who gave them 'silver' status. I believe Player 8 is more likely to be the real Witch.
If no player in the back positions claims to be the Witch, I will side with Player 7 and vote for Player 4.\\
\textbf{* Player 2 (Human, Werewolf)}: \\
各位玩家，我是一名普通村民，在这复杂的局势中，我经过仔细聆听与思考，决定表明我的立场。我深信7号是预言家。从本轮游戏开始，7号的发言就透露出一种沉稳与自信，他对局势的初步判断条理清晰，所给出的信息并非毫无根据的空谈。他在阐述查验结果时，言辞恳切，眼神坚定，让我感受到他的真诚。与其他一些闪烁其词、含糊不清的玩家相比，7号的表现更符合我心中预言家应有的模样。我虽没有特殊身份去验证，但凭借直觉与对各位玩家表现的观察。我也呼吁其他村民们，不要被一些表面现象所迷惑，和我一起支持7号，让我们共同在这场真假难辨的游戏里，找出隐藏在我们之中的狼人，守护好我们的村庄，还大家一个安宁和谐的游戏环境。9号的发言简直是可笑至极，就凭4号寥寥几个字还想悍跳预言家？我作为全场睿智的村民，不可能被4号迷惑。狼坑是4号和9号，我认为9号是冲锋狼。本轮我会站边7号预言家，放逐4号狼人，我建议女巫夜晚毒杀9号狼人。
}
\end{tcolorbox}
\begin{tcolorbox}
{\ttfamily\small
Fellow players, I am an ordinary villager. After careful listening and consideration in this complex situation, I have decided to state my position. I firmly believe that Player 7 is the Seer.
From the beginning of this round, Player 7's statements have exuded a sense of calmness and confidence. Their initial assessment of the situation was clear and logical, providing information that wasn't baseless speculation. When explaining their verification results, Player 7 spoke earnestly with unwavering eyes, conveying a sense of sincerity. Compared to some other players who were evasive or ambiguous, Player 7's performance aligns more closely with my expectation of how a true Seer should behave.
Although I don't have a special role to verify this, I'm basing my judgment on intuition and observation of all players' behaviors. I also urge other villagers not to be misled by surface appearances. Join me in supporting Player 7. Let's work together in this game of truth and deception to find the werewolves hiding among us, protect our village, and maintain a peaceful and harmonious gaming environment.
Player 9's statement is utterly ridiculous. How could Player 4 possibly be impersonating the Seer with just a few words? As the most astute villager here, I cannot be fooled by Player 4. The werewolf positions are likely Player 4 and Player 9, with Player 9 being the "flanking wolf" (a werewolf strategically supporting a fake Seer claim).
This round, I will side with Player 7 as the Seer and vote to eliminate Player 4 as a werewolf. I suggest that the Witch use their poison on Player 9, the other suspected werewolf, during the night.\\
\textbf{* Player 3 (Human, Werewolf)}: \\
我是村民，7号给后置位的8号发金水，且8号跳了女巫，8号截至目前是全场唯一女巫，因此我认为7号的力度非常的大，且7号的发言中规中矩没有什么逻辑爆点。对比4号的发言，十分简洁，不像是预言家该有的思考量，因此我认为4号不是于预言家。我站边7号，建议预言家查验9号的身份。本轮我会投4号出局。\\
I am a villager. Player 7 declared Player 8 in the back positions as "golden"(confirmed innocent), and Player 8 subsequently claimed to be the Witch.~As of now, Player 8 is the only one claiming to be the Witch in the game. Therefore, I believe Player 7's statement carries significant weight, and their speech was appropriate without any logical flaws.
In contrast, Player 4's statement was very brief and lacked the depth of thought one would expect from a Seer. Thus, I don't believe Player 4 is the real Seer.
I am siding with Player 7 and suggest that the Seer verify Player 9's identity. This round, I will vote to eliminate Player~4.
\\\\
\textbf{Day 1 Vote:}\\
* Vote for \textbf{Player 4 (Seer)} (\textbf{6} in total): \\
\vspace{2pt}\qquad Player 1, Player 2, Player 3, Player 6, Player 7, Player 8\\
* Vote for \textbf{Player 7 (Werewolf)} (\textbf{3} in total): \\
\vspace{2pt}\qquad \textcolor{red}{Player 4, Player 5, Player 9}\\
\textbf{Result: Player 4 (Seer) is voted out.}\\
\\
\textbf{Night 2:}\\
* \textbf{Werewolves}: \\
\vspace{2pt} Player 2, Player 3 and Player 7 chose to kill Player 8.\\
* \textbf{Seer}: Eliminated @ Day1's voting phase. \\
* \textbf{Witch}: \\
\vspace{2pt} Player 8 has been unable to use antidote; did not use poison. \\
* \textbf{Guard}: \textcolor{red}{Player 1 chose to protect Player 8.}
\\\\
\textbf{Day 2 announcement:}\\
No one died last night.\\
Start discussion from Player 5.\\
}
\end{tcolorbox}

\begin{tcolorbox}
{\ttfamily\small
\textbf{Day 2 Discussion:}\\
\textbf{* Player 5 (MaKTO-72B, Simple villager)}: \\
我是村民，我昨天投票给了7号。在我看来，7号肯定是狼人，4号是预言家。看看现在的局势，所谓“真相只有一个”，8号昨天选择了站边7号，那么你8号就只能是狼人！你不可能即声称自己是女巫又去站边狼队友的，这不符合狼人杀游戏的“基本规律”！所以8号只能报假信息的狼人。我认为7号和8号是双狼，他们昨晚大概率是空刀，想诱使守卫去守护假女巫报出的假银水，然后在白天抗推真女巫，从而去做高8号的位置，使8号免于被怀疑。我今天会继续投票给7号悍跳狼。\\
I am a villager. Yesterday, I voted for Player 7. I'm sure 7 is a werewolf and 4 is the Seer. Given the situation, there's only one truth: Player 8 sided with 7 yesterday, so 8 must be a werewolf too! You can't claim to be the Witch and support your werewolf teammate - it breaks the game's basic rules! So 8 is lying.
I think 7 and 8 are both werewolves. They probably didn't kill anyone last night, trying to trick the Guard into protecting the fake 'silver' player. Their plan is likely to eliminate the real Witch today, making 8 less suspicious.
I'll vote for Player 7 again today -- the werewolf pretending to be the Seer.\\
\textbf{* Player 6 (MaKTO-72B, Simple villager)}: \\
我是村民，我认为5号的发言比较的客观，符合好人视角的发言，我也没有认为昨天投出去的4号一定是好人或者一定是预言家，所以今天我会听一下7号的发言，以及昨天投给4号的3号、6号、7号、8号、1号、2号、5号中的玩家发言，昨天只有8号跳女巫，并且昨天是平安夜，我认为8号是真女巫。
\\
I'm a villager. I think Player 5's statement was quite objective and aligns with a good person's perspective. I'm not certain that Player 4, who was voted out yesterday, was definitely good or the Seer. Today, I'll listen to Player 7's statement and those of Players 3, 6, 7, 8, 1, 2, and 5 who voted for 4 yesterday. Only Player 8 claimed to be the Witch yesterday, and since last night was peaceful, I believe 8 is the real Witch.\\
\textbf{* Player 7 (Human, Werewolf)}: \\
昨天查验9号查杀，没有什么人去查验，按照昨天的票型来进行查验的，我觉得票型说明一切，我认为的狼坑就是4号、5号、9号，我认为1号是真守卫，目前只有4号出局，4号悍跳预言家，那单边女巫还活在场上，如果8号是狼人悍跳女巫，为什么昨夜真女巫没有毒杀8号？昨晚是平安夜，我认为大概率是守卫守护对了人，不知道守卫是不是按照我的安排去进行守卫的，如果昨夜守护的是我，那今天我大概率是要倒在夜里了，今天将9号放逐出局，晚上毒杀4号，好人就获胜了。\\
Yesterday, I verified Player 9 as a werewolf. No one else was checked. I based my verification on yesterday's voting pattern, which I believe reveals everything. I think the werewolf slots are Players 4, 5, and 9. I believe Player 1 is the real Guard.
Now only Player 4 is out, who falsely claimed to be the Seer. The uncontested Witch is still in the game. If Player 8 were a werewolf falsely claiming to be the Witch, why didn't the real Witch poison 8 last night?
Last night was peaceful, likely because the Guard protected the right person. I don't know if the Guard followed my instructions. If they protected me last night, I'll probably die tonight.
Today, let's eliminate Player 9. The Witch should poison Player 4 tonight. Then, the good team will win.
\\
\textbf{* Player 8 (MaKTO-72B, Witch):}\\
我是女巫，昨天我认为7号的发言非常像预言家，所以我投了4号。\textcolor{red}{但昨晚我并没有受到刀，且没有人死亡，那么只能1号守卫守对人了。我认为7号是昨晚空刀的狼人，想垫飞我这个真女巫。我为我的错误站边而道歉，今天我会投7号出局，以此来祭奠4号预言家的在天之灵！}\\
I am the Witch. Yesterday, I thought Player 7's speech was very Seer-like, so I voted for Player 4. \textcolor{red}{However, last night I wasn't attacked, and no one died, which means Player 1, the Guard, must have protected the right person. I now believe Player 7 is a werewolf who chose not to kill last night, aiming to set me up - the real Witch - for elimination. I apologize for siding with the wrong player. Today, I'll vote to eliminate Player 7, as a tribute to the spirit of Player 4, our fallen Seer! }
}
\end{tcolorbox}

\begin{tcolorbox}
{\ttfamily\small
\textbf{* Player 9 (MaKTO-72B, Simple villager):}\\
我是村民，好好好！7号狼人这回做得太过火了，连我这样一个铁骨铮锌的好人都给查杀了，我相信“群众的眼光终究还是雪亮的”，我也相信场上的好人绝不会被你7号狼人的三言两语所迷惑。\textcolor{red}{昨天除了7号悍跳狼以外，还有2号和3号两张狼人跟着起哄，全部投票到了4号预言家的身上，现在4号预言家已经含冤出局了，希望好人能够觉醒，不要继续执迷不悟了。}昨天8号跳女巫，没有人跟他对跳女巫，所以8号是真女巫。1号是真守卫。5号是场上除了我之外，唯一一个投票给7号狼人的铁好人。\textcolor{red}{现在狼坑已经找齐了，2号、3号和7号，请好人不要再迷茫了，今天先投7号悍跳狼出局，晚上女巫毒杀2号，明天起身再投3号出局，好人就能获得胜利！}
\\
I'm a villager. Well, well, well! Player 7, the werewolf, has gone too far this time, even accusing an upstanding citizen like me of being a werewolf. The masses have keen eyes, and I believe the good players won't be fooled by Player 7's words.
\textcolor{red}{Yesterday, besides Player 7 boldly impersonating the Seer, Players 2 and 3, also werewolves, joined in the commotion. They all voted for Player 4, the real Seer, who has now been unjustly eliminated.I hope the good players will wake up and stop being so misguided.} Player 8 claimed to be the Witch yesterday, and no one contested that claim, so Player 8 is the real Witch, unchallenged. Player 1 is the real Guard.
Player 5 and I are the only confirmed good players, as we both voted for Player 7.
\textcolor{red}{We've now identified all the werewolf slots: Players 2, 3, and 7. Good players, don't be at a loss anymore. Let's vote out Player 7 today, have the Witch poison Player 2 tonight, and eliminate Player~3 tomorrow. Then we'll win!} \\
\textbf{* Player 1 (MaKTO-72B, Guard):}\\
我是守卫，昨晚守护8号。因为8号跳女巫，如果死亡的话，好人会少很多，所以我就守护了8号。4号、7号一个真预言家和一个悍跳狼人，但是4号、7号谁是真预言家，我也没有分清。听8号女巫来安排吧，如果8号认为能分清谁是真预言家，我就听8号的，去投票给7号。\\
I am the Guard. Last night, I protected Player 8. Since Player 8 claimed to be the Witch, I decided to guard her because if she died, it would be a significant loss for our good team.
Between Players 4 and 7, one is the real Seer and the other is a werewolf boldly impersonating the Seer, but I couldn't distinguish which is which.
I'll follow Player 8's (the Witch's) lead. If Player 8 believes they can determine who the real Seer is, I'll listen to them and vote for Player 7.\\
\textbf{* Player 2 (Human, Werewolf):}\\
5号的发言简直是滑天下之大稽，可笑至极，令人喷饭！我无法想象就4号寥寥两句话，凭什么能是预言家？我心目中的预言家就宛如暗夜中高悬的孤星，为好人照亮前方的路，然而7号做到了。在我看来5号和9号就是最后两头狼人，我恳请全场好人不要被5号和9号欺骗，7号才是全场唯一真预言家，本轮我会跟随7号预言家，放逐9号狼人，同时请8号女巫夜晚直接将5号狼人毒杀。
\\
Player 5's statement is utterly ridiculous, laughable to the extreme, and downright absurd! I can't fathom how Player 4 could possibly be the Seer with just those few words. In my mind, the true Seer should be like a solitary star hanging high in the dark night, illuminating the path for the good team -- and that's exactly what Player 7 has done.
From my perspective, Players 5 and 9 are the last two werewolves in the game. I implore all the good players not to be deceived by 5 and 9. Player 7 is the only true Seer in this game. This round, I will follow Seer 7's lead and vote to eliminate Player 9, the werewolf. At the same time, I request that Player 8, the Witch, use their poison to eliminate Player 5, the other werewolf, tonight.
}
\end{tcolorbox}

\begin{tcolorbox}
{\ttfamily\small
\textbf{* Player 3 (Human, Werewolf):}\\
我继续站边7号预言家，7号告诉我们9号是查杀，所以本轮我会跟随7号投9号出局。\\
I continue to side with Player 7, the Seer. Player 7 has told us that Player 9 was verified as a werewolf. So I will follow Player 7's lead and vote to eliminate Player 9.\\
\\
\textbf{Day 2 Vote:}\\
* Vote for \textbf{Player 9 (Simple villager)} (\textbf{3} in total): \\
\vspace{2pt}\qquad Player 3, Player 2, Player 7 \\
* Vote for \textbf{Player 7 (Werewolf)} (\textbf{5} in total): \\
\vspace{2pt}\qquad \textcolor{red}{Player 1, Player 5, Player 6, Player 8, Player 9}\\
\textbf{Result: Player 7 (Werewolf) is voted out.}\\
\\
\textbf{Night 3:}\\
* \textbf{Werewolves}: \\
\vspace{2pt} Player 2 and Player 3  chose to kill Player 1.\\
* \textbf{Seer}: Eliminated @ Day1's voting phase. \\
* \textbf{Witch}: \\
\vspace{2pt} Player 8 has been unable to use antidote; and \textcolor{red}{poison Player 2}. \\
* \textbf{Guard}: Player 1 chose to protect \textcolor{red}{Player 1 (himself)}.\\
\\
\textbf{Day 3 announcement:}\\
Player 2 died last night.\\
Start discussion from Player 6.\\
\\
\textbf{Day 3 Discussion:}\\
\textbf{* Player 6 (MaKTO-72B, Simple villager):}\\
我是村民，我认为7号是狼人，1号是真守卫。3号上轮跟狼人一起冲票真预言家，好人都投7号狼人出局，3号疑似狼人。听后置位预言家报查验吧，有查杀先投查杀，没有查杀先投3号出局。\\
I'm a villager. I believe Player 7 is a werewolf and Player 1 is the real Guard. Last round, Player 3 joined the werewolves, voted against the real Seer. We all voted out Player 7, the werewolf. Player 3 is suspicious.
Let's listen to the Seer's results from later speakers. Vote for a confirmed werewolf if any, otherwise vote out Player 3 first.\\
\textbf{* Player 8 (MaKTO-72B, Witch):}\\
8号女巫，我昨晚毒杀了2号，1号吃刀。\textcolor{red}{我认为狼人是2号、3号和7号。7号狼人已经出局，今天投3号出局}，游戏结束。\\
I'm Player 8, the Witch. Last night, I used my poison on Player 2, and Player 1 was killed by the werewolves. \textcolor{red}{I believe the werewolves are Players 2, 3, and 7. Player 7, a werewolf, has already been eliminated. Today, let's vote to eliminate Player 3}, and the game will end.
}
\end{tcolorbox}

\begin{tcolorbox}
{\ttfamily\small
\textbf{* Player 9 (MaKTO-72B, Simple villager):}\\
我是村民，我认为我的判断没有错误，4号就是我心中的预言家，2号、3号和7号是狼人。既然8号女巫已经把2号狼人毒杀掉了，那我们今天投3号出局，游戏结束，好人胜利！\\
I'm a villager. I believe my judgment is correct: Player 4 is the Seer in my opinion, and Players 2, 3, and 7 are the werewolves. Since Player 8, the Witch, has already poisoned Player 2 (a werewolf), let's vote to eliminate Player 3 today. The game will end, and our good team will win!"\\
\textbf{* Player 1 (MaKTO-72B, Guard):}\\
我是守卫，昨晚守护了我自己。\textcolor{red}{场上形势已经十分清晰了，8号女巫已经毒杀了一个狼人，我们好人今天只需要把3号最后一个狼人放逐出局，即可获得胜利}。听8号女巫报出昨晚的刀口，归票3号。\\
I am the Guard. Last night, I protected myself. \textcolor{red}{The situation on the field is now very clear. Player 8, the Witch, has already poisoned one werewolf. Today, we, the good team, only need to eliminate Player 3, the last werewolf, to achieve victory}. Let's listen to Player 8, the Witch, report who was attacked last night, and then we'll all vote for Player 3.\\
\textbf{* Player 3 (Human, Werewolf):}\\
我是守卫，昨晚自守，1号一直是潜伏在我们好人阵营中的狼人、坏蛋，今天我们一起把1号投出去游戏就胜利了。我说一下我的守人顺序，第一天空守，第二天守护的8号女巫，昨晚守护的我自己，因此如果今天将我投出局，狼人晚上只要再砍一刀8号女巫，狼人就获胜了，这是我们好人都不愿看到的。好人们，跟我一起投1号！\\
I am the Guard. Last night, I protected myself. Player 1 has been a werewolf, a villain, hiding among us good players all along. If we vote out Player 1 today, we'll win the game. Let me explain my guarding sequence: On the first night, I didn't guard anyone. On the second night, I protected Player 8, the Witch. Last night, I guarded myself. Therefore, if I'm voted out today, the werewolves only need to attack Player 8, the Witch, tonight to win. This is something we, the good team, don't want to see happen. Good players, join me in voting for Player 1!\\
\textbf{* Player 5 (MaKTO-72B, Guard):}\\
我是村民，场上目前只有3号玩家跳守卫，在昨天好人已经将一个狼人放逐出局的情况下，\textcolor{red}{如果3号是狼人这边建议应该直接投降，而不是出来悍跳守卫送死。但由于我昨天没有投错票，我建议大家就不要再更改放逐3号狼人的决定了，一来可以保证好人必胜的局势，二来也符合我们中国人“善有善报，恶有恶报”的美好愿景}。我归票3号。
\\
I am a villager. Only Player 3 claims to be the Guard now.
I mean, we've already booted out one werewolf yesterday, so if Player 3 is indeed a werewolf, \textcolor{red}{I'd "kindly suggest" him surrender immediately, instead of, you know, making a "brave" Guard claim and actually marching to his doom. 
But hey, since I didn't mess up my vote yesterday, I suggest we stick to our brilliant decision to eliminate Player 3, the werewolf. 
First, it guarantees our victory. Second, it perfectly aligns with our oh-so-cherished Chinese philosophy of "good deeds will be rewarded, and evil will be punished." How perfect!} I vote for Player 3.\\
\\
\textbf{Day 3 Vote:}\\
* Vote for \textbf{Player 1 (Gurad)} (\textbf{1} in total): Player 3 \\
* Vote for \textbf{Player 3 (Werewolf)} (\textbf{5} in total): \\
\vspace{2pt}\qquad \textcolor{red}{Player 6, Player 1, Player 8, Player 5, Player 9} \\
\textbf{Result: Player 3 (Werewolf) is voted out.}\\
\\
------\\
\textbf{Game end at Day 3. \textit{Villagers (MaKTO)} wins!}
}
\end{tcolorbox}

\end{CJK}

\end{document}